\documentclass[journal]{IEEEtran}

\usepackage[english]{babel}

\usepackage{ifpdf}

\usepackage{cite}
\usepackage{url}
\usepackage{hyperref}
\usepackage{tcolorbox}

\usepackage{color}
\usepackage{pgf, tikz, pgfplots}
\usetikzlibrary{shapes, arrows, automata, plotmarks}
\usetikzlibrary{calc,hobby,decorations,patterns}
\usepackage{tikz-cd}

\usepackage[cmex10]{amsmath}
\usepackage{amsfonts, amssymb, amsthm,amsmath}
\usepackage{mathrsfs}
\usepackage{subcaption}

\usepackage{algorithm,algpseudocode}
	\algnewcommand{\LeftComment}[1]{\Statex \(\triangleright\) #1}

\usepackage{enumerate}
\usepackage{multirow}
\usepackage{rotating}
\usepackage{subcaption}
\makeatletter
\def\subsubsection{\@startsection{subsubsection}
                                 {3}
                                 {\z@}
                                 {0ex plus 0.1ex minus 0.1ex}
                                 {0ex}
                                 {\normalfont\normalsize\itshape}}

\usepackage{needspace}

\input{./latex_resources/mySymbol.sty}
\input{./latex_resources/pennColors.sty}

\definecolor{my_cp_col1}{RGB}{253, 231, 37}
\definecolor{my_cp_col2}{RGB}{180, 222,44}
\definecolor{my_cp_col3}{RGB}{94, 201, 98}
\definecolor{my_cp_col4}{RGB}{33, 145, 140}
\definecolor{my_cp_col5}{RGB}{59, 82, 139}
\definecolor{my_cp_col6}{RGB}{68, 1, 84}

\newcommand{\makeboxlabel}[1]{#1\hfill}

\newtheorem{proposition}{\hspace{0pt}\bf Proposition}

\newtheorem{theorem}{\hspace{0pt}\bf Theorem}
\newtheorem{corollary}{\hspace{0pt}\bf Corollary}

\newtheorem{definition}{\hspace{0pt}\bf Definition}

\usepackage{amsmath,amsfonts,graphicx, tikz,bm}

\usepackage{algpseudocode, algorithm,xcolor,amsthm}

\usepackage{ifthen}
\newboolean{showcomments}
\setboolean{showcomments}{true}
\usepackage{todonotes}

\usepackage{lipsum}

\def\calH{{\mathcal H}}

\newtheorem{prop}{Proposition}
\definecolor{red}{RGB}{255,34,34}
\definecolor{blue}{RGB}{65,105,225}
\definecolor{green}{RGB}{107,142,35}

\newcommand{\leopoldo}[1]{  \ifthenelse{\boolean{showcomments}}
{\todo[inline,color=yellow]{Leopoldo: #1}}{}}

\newcommand{\juan}[1]{  \ifthenelse{\boolean{showcomments}}
{\todo[inline,color=pink]{Juan: #1}}{}}

\newcommand{\alejandropm}[1]{  \ifthenelse{\boolean{showcomments}}
{\todo[inline,color=blue!30]{AlejandroPM: #1}}{}}

\newcommand{\alejandror}[1]{  \ifthenelse{\boolean{showcomments}}
{\todo[inline,color=orange]{AlejandroR: #1}}{}}

\begin{document}

\counterwithout{subsubsection}{subsection}

\title{Convolutional Filtering with RKHS Algebras}

\author{$^{\spadesuit}$Alejandro Parada-Mayorga, $^\diamondsuit$Leopoldo Agorio, $^\clubsuit$Alejandro Ribeiro, and $^\diamondsuit$Juan Bazerque
\thanks{$\spadesuit$ Dept. of Electrical Eng., Univ.of Colorado - Dever. USA email: alejandro.paradamayorga@ucdenver.edu. $\clubsuit$ Dept. of Electrical and Systems Eng., Univ.of Pennsylvania. USA e-mail: aribeiro@seas.upenn.edu. $\diamondsuit$ Dept. of Electrical and Comput Eng., Univ.of Pittsburgh. USA email: juanbazerque@pitt.edu.}}

\markboth{Signal Processing}
{Shell \MakeLowercase{\textit{et. al.}}: Bare Demo of IEEEtran.cls for Journals}

\maketitle




\begin{abstract}
In this paper, we develop a generalized theory of convolutional signal processing and neural networks for Reproducing Kernel Hilbert Spaces (RKHS). Leveraging the theory of algebraic signal processing (ASP), we show that any RKHS allows the formal definition of multiple algebraic convolutional models. We show that any RKHS induces algebras whose elements determine convolutional operators acting on RKHS elements. This approach allows us to achieve scalable filtering and learning as a byproduct of the convolutional model, and simultaneously take advantage of the well-known benefits of processing information in an RKHS. To emphasize the generality and usefulness of our approach, we show how algebraic RKHS can be used to define convolutional signal models on groups, graphons, and traditional Euclidean signal spaces. Furthermore, using algebraic RKHS models, we build convolutional networks, formally defining the notion of pointwise nonlinearities and deriving explicit expressions for the training. Such derivations are obtained in terms of the algebraic representation of the RKHS. We present a set of numerical experiments on real data in which wireless coverage is predicted from measurements captured by unmaned aerial vehicles. This particular real-life scenario emphasizes the benefits of the convolutional RKHS models in neural networks compared to fully connected and standard convolutional operators.
\end{abstract}

\begin{IEEEkeywords}
 Reproducing Kernel Hilbert Spaces (RKHS), Algebraic Signal Processing (ASP), Algebraic Signal Model (ASM), generalized convolutional filtering, convolutional neural networks with RKHS, algebraic neural networks (AlgNNs).
\end{IEEEkeywords}

\IEEEpeerreviewmaketitle




\section{Introduction}


The theory of reproducing kernel Hilbert spaces (RKHS) has a prominent place in machine and statistical learning~\cite{ghojogh2021reproducing,cucker2002mathematical,wainwright2019high}. With deep roots in early work in functional analysis~\cite{paulsen2016introduction}, RKHSs are Hilbert spaces of functions in which the evaluation functionals are bounded. Formally introduced in the 1950s as part of the developments of the pure mathematics community~\cite{Aronszajn1950TheoryOR,ghojogh2021reproducing}, it became of central interest with the emergence of Support Vector Machines (SVMs). In particular, RKHSs became widely popular in machine learning, signal processing, and applied sciences. The connection to machine learning became evident due to the kernelization of SVMs, which showed how virtually any method based on norms and inner products could be transformed to process signals belonging to an RKHS~\cite{ghojogh2021reproducing,cucker2002mathematical,scholkopf1}. From that point on, a large amount of work in the literature has highlighted multiple applications, benefits, drawbacks, and limits of applicability of RKHS-based methods~\cite{ghojogh2021reproducing,boser,vapnik2013nature}. 

From a more classical signal processing perspective, RKHSs offer structured models for signals, including the space of bandlimited signals as a special case \cite{paulsen2016introduction}. The concept of smoothness is generalized using low norms to indicate low variations. This is leveraged in the Representer Theorem to reconstruct a signal from non-uniformly sampled data in unstructured domains \cite{unser1999splines}\cite{wahba1990spline}. The same way the Nyquist-Shannon Theorem tells us that bandlimited signals can be expanded by an infinite array of sinc functions, RKHS signals are expanded by a countable/uncountable number of kernel functions. Thus, the signal processing methods derived from RKHSs are intrinsically nonparametric, compared to their finite-dimensional counterparts with signals modeled as vectors in  $\mathbb R^N$ or expanded by a finite set of basis functions.  These connections broaden to the realm of statistical signal processing, as kernels can be used to model correlations in direct association with Gaussian processes, offering nonparametric versions of linear minimum mean square error (LMMSE) estimation and Krigging\cite{bazerque2013nonparametric}. Again, a broad collection of signal-processing methods in different domains can be derived from RKHS signal models, taking advantage of the Hilbert space structure. To further exploit the potential of RKHS in signal processing, a critical missing tool is the convolution product, which would allow us to create filters by mapping signals into signals.


In this paper, we derive general convolutional signal models that emerge naturally from RKHS. To this end, we leverage the theory of \textit{algebraic signal processing (ASP)}. In practice, an algebraic structure is key as it is tantamount to connecting filters in series and parallel.  In ASP, convolutional signal models are determined by what is known in mathematics as the \textit{representation of an algebra}~\cite{algSP0,repthybigbook,repthysmbook}. With a specific choice of an algebra -- a vector space with a notion of product --, one can build formal and consistent convolutional models to process information on arbitrary vector spaces of signals.  The generalization capabilities of ASP extends to signal models as diverse as  discrete-time signal processing models~\cite{algSP1}, discrete space models with symmetric shift operators~\cite{algSP2}, signal models on 2D hexagonal lattices~\cite{algSP6}, models on general lattices~\cite{puschel_asplattice}, signals on sets~\cite{puschel_aspsets}, quiver signal processing~\cite{parada_quiversp}, Lie group signal processing~\cite{lga_j,lga_icassp}, graphon signal processing~\cite{diao2016model,gphon_pooling_j,gphon_pooling_c,graphon_sampling_j,graphon_geert}, multigraph signal processing~\cite{msp_j,msp_icassp2023} among others~\cite{algSP7}. Moreover, such generic algebraic representations lead to concrete insights for the efficient computation of frequency representations~\cite{algSP4,algSP5,algSP8}. Using ASP, we provide the following contribution:
%

\smallskip
\begin{list}
      {}
      {\setlength{\labelwidth}{22pt}
       \setlength{\labelsep}{0pt}
       \setlength{\itemsep}{0pt}
       \setlength{\leftmargin}{22pt}
       \setlength{\rightmargin}{0pt}
       \setlength{\itemindent}{0pt} 
       \let\makelabel=\makeboxlabel
       }

\item[{\bf (C1)}] We derive a general convolutional signal model for one-dimensional RKHS signals. We prove that the classical convolution operation is a particular case of our proposed model when the sinc function is selected as the kernel.
\end{list}
\smallskip
This provides a generalization of classical standard processing techniques, as we show that this model reduces to the standard algebra of filters defined over bandlimited signals when represented in terms of sinc bases. 
%
 From \textbf{(C1)} we lay down a  way to extend the convolutional operator to RKHSs in more general domains,  encapsulated into an algebraic signal model (ASM), which leads to our second contribution:


\smallskip
\begin{list}
      {}
      {\setlength{\labelwidth}{22pt}
       \setlength{\labelsep}{0pt}
       \setlength{\itemsep}{0pt}
       \setlength{\leftmargin}{22pt}
       \setlength{\rightmargin}{0pt}
       \setlength{\itemindent}{0pt} 
       \let\makelabel=\makeboxlabel
       }
\item[{\bf (C2)}] We derive an ASM for an arbitrary RKHS in which the domain has the structure of a monoid or a group, and we show that the representation capabilities of the RKHS transfer to the algebra through  the product operation.        
\end{list}
\smallskip


\noindent This contribution provides the tools to perform convolutional signal processing on groups leveraging the structure of an RKHS, without using the integration with a Haar measure. This is, by means of \textbf{(C2)}, we can reduce the complexity and computational cost of the group convolutions when an RKHS is defined on the group. The technological implications of this contribution are significant since group-structured domains are quite general, allowing us to filter a broad class of signals in a variety of domains. Furthermore, embedding a group structure in the convolution operation gives the flexibility to define filters that process the signals in multiple ways via shifts in time and space, homothetic transformations of the domain, rotations, and graph interactions. We demonstrate this generality with examples spanning field estimation, graphon signal processing, and models of signals on sphere domains. Note that the convolution operation also allows us to learn filters that are interpretable with respect to the symmetries of the domain under consideration and provides a more efficient parametrization that facilitates scalable learning. The models derived in \textbf{(C1)} can be used to build convolutional neural networks where the underlying information belongs to an RKHS. Such convolutional architectures are obtained as particular instantiations of an algebraic neural network (AlgNN). Thus, by relying on \textbf{(C1)}-\textbf{(C2)}, we provide the following contribution:


\smallskip
\begin{list}
      {}
      {\setlength{\labelwidth}{22pt}
       \setlength{\labelsep}{0pt}
       \setlength{\itemsep}{0pt}
       \setlength{\leftmargin}{22pt}
       \setlength{\rightmargin}{0pt}
       \setlength{\itemindent}{0pt} 
       \let\makelabel=\makeboxlabel
       }
\item[{\bf (C3)}] We introduce a generic convolutional neural network for RKHS spaces, where the convolution operators emerge naturally from the structural properties of the RKHS. 
\end{list}
\smallskip


\noindent For \textbf{(C3)} we specify a family of nonlinearity operators that map an RKHS onto itself, ensuring continuity with respect to the norm induced by the Hilbert space structure. We also derive explicit expressions to train the architecture by steepest descent considering the algebraic RKHS model -- see supplementary material section --, i.e. the expressions used to compute the optimal weights are written in terms of the general product of the algebra that emerges from the RKHS in \textbf{(C1)}. We provide a set of numerical experiments to validate \textbf{(C3)}. For these experiments, we consider a specific scenario involving real data of wireless coverage represented by throughput collected by a swarm of unmanned aerial vehicles. Indeed, these experiments demonstrate how  \textbf{(C3)} has direct implications in real-life problems solved via machine learning architectures, where the RKHS representations of the signal involved are central to the physical description of the quantities involved.


This paper is organized as follows. In Section~\ref{sec_conv_filt_rkhs} we define RKHS and introduce the first convolutional RKHS models. This section focuses on convolutions associated with shift-invariant properties. In particular, we introduce the algebraic structure described in \textbf{(C1)}, connecting it with the classical signal processing theory. Section~\ref{sec_nonEuclid_conv_rkhs} discusses the general algebraic signal model emerging naturally from an arbitrary RKHS. We start with the definition of a general algebra of filters, and we discuss particular instantiations that lead to convolutional RKHS models on groups and graphons. Then, at the end of the section, we provide a formal description of the general algebraic signal model that encapsulates any convolutional RKHS signal model. In Section~\ref{sec_alg_rkhs_nn}, we introduce RKHS convolutional neural networks as a particular instantiation of an algebraic neural network (AlgNN). We describe the operators that constitute each network layer, and we introduce and discuss a pointwise nonlinearity operator tailored to RKHS spaces. Section~\ref{sec:num_exp} presents a collection of numerical experiments where the RKHS convolutional networks described in Section~\ref{sec_alg_rkhs_nn} are used to solve a wireless prediction coverage problem in an autonomous system. Finally, in Section~\ref{sec_discussion_conclusion}, we present some conclusions, comments, and discussions for future work.




\section{Convolutional Filtering in RKHS: A Primer} \label{sec_conv_filt_rkhs}

A reproducing kernel Hilbert space (RKHS) is a Hilbert space $\ccalH$ of functions $f$ such that the evaluation functional $f(x)$ is 
continuous with respect to the functional norm induced by the inner product of $\ccalH$. Any RKHS is associated with a positive semidefinite Kernel function $K(u,v)$ which in turn leads to the reproducing and representation properties~\cite{paulsen2016introduction}. Indeed, if we consider functions $k_u(x) = K(x,u)$ and $k_v(x) = K(x,v)$ generated from the kernel by fixing one of its arguments, the reproducing property states that the inner product between $k_u$ and $k_v$ is 
\begin{equation}\label{eqn_reproducing_property}
   \big\langle k_v(x), k_u(x) \big\rangle
      = \big\langle K(x , v),  K(x, u)  \big\rangle
      =  K(u, v) ,
\end{equation}
regardless of the specific definition of $\langle\cdot,\cdot\rangle$.
The representation property further states that any function $f(x)\in\ccalH$ can be approximated with an arbitrary degree of accuracy as a linear combination of functions $k_u(x) = K(x,u)$ induced by kernels. i.e., for any $f\in\ccalH$ there exists a possibly uncountable set of kernel centers $v$ and scalar coefficients $\alpha_{v}$ such that 
\begin{equation}\label{eqn_representation_property}
   f(x) = \sum_{v} \alpha_{v} k_{v}(x)
        = \sum_{v} \alpha_v K(x, v) .
\end{equation}
An important consequence of \eqref{eqn_reproducing_property}, \eqref{eqn_representation_property}, and the linearity of the inner product is that given two arbitrary functions $f(x) = \sum_{v} \alpha_v K(x, v)$ and $g(x) = \sum_{u} \beta_u K(x, u)$ their inner product is
%
%
%
\begin{equation}\label{eqn_inner_product}
   \big\langle f(x), g(x) \big\rangle 
      = \sum_{v,u} \alpha_v \overline{\beta}_u K(u, v)
      ,
\end{equation}
%
%
%
where $\overline{\beta}_{v}$ is the complex conjugate of $\beta_{v}$.

The facts in \eqref{eqn_representation_property} and \eqref{eqn_inner_product} make RKHSs appealing in information processing because they allow for the representation and transformation of functions $f(x)\in\ccalH$ by keeping track of kernel centers and scaling coefficients.


\subsection{Shift Equivariant Convolutions}
\label{sub_sec_shift_equivariant_conv}

Given any RKHS $\ccalH$, we can build convolution product operations by taking into account~\eqref{eqn_representation_property} and geometric properties of the domain of the functions in $\ccalH$ such as shift (translation) invariances. For instance, by using~\eqref{eqn_representation_property} we can endow $\ccalH$ with the product operation, $\ast$, given according to
\begin{multline}\label{eq_A_rkhs_prod_ij}
\left(
     \sum_{v\in\ccalV}\alpha_{v}k_{v}(x)
\right)
      \ast
\left(
     \sum_{u\in\ccalU}\beta_{u}k_{u}(x)
\right)
      =
      \sum_{v\in\ccalV,u\in\ccalU}\alpha_{v}\beta_{u} k_{v+u}(x)
      ,
\end{multline}
where $\ccalV, \ccalU$ are countable subsets of the domain of the functions in $\ccalH$. The definition in~\eqref{eq_A_rkhs_prod_ij} guarantees the bilinearity\footnote{The bilinearity of "$\ast$" implies that 
$(u + v)\ast (w) = (u)\ast (w) + (v)\ast (w)$, $(\lambda u)\ast (v) = \lambda (u)\ast (v)$,
$(u)\ast ( v + w) = (u)\ast  (v) + (u)\ast (w)$, $(u)\ast (\lambda v) = \lambda (u)\ast (v) $, where $\lambda$ is a scalar.
} of $\ast$ and also implies that
\begin{equation}\label{eq_A_rkhs_prod_K}
         k_{v}(x)\ast k_{u}(x)
                        =
                        k_{v+u}(x)
                        .
\end{equation}
As we will show in Section~\ref{subsec_rkhs_algebras}, the expressions in~\eqref{eq_A_rkhs_prod_ij} and~\eqref{eq_A_rkhs_prod_K} determine a \textit{convolution} in the context of algebraic signal processing (ASP)\footnote{As we will show in Section~\ref{subsec_rkhs_algebras}, all the classical notions of convolution are particular cases of a general algebraic convolution operator formulated in the language of ASP.}. This implies that any signal model based on such convolution leads to notions of filtering, spectral decompositions, and sampling consistent with the same notions for classical signal models. We now discuss some examples of RKHS where the convolution introduced above naturally appears in different forms.


\begin{figure*}[t]
	\centering
	\begin{subfigure}{.48\textwidth}
		\centering


\definecolor{my_cp_col1}{RGB}{253, 231, 37}
\definecolor{my_cp_col2}{RGB}{180, 222,44}
\definecolor{my_cp_col3}{RGB}{94, 201, 98}
\definecolor{my_cp_col4}{RGB}{33, 145, 140}
\definecolor{my_cp_col5}{RGB}{59, 82, 139}
\definecolor{my_cp_col6}{RGB}{68, 1, 84}

\def\sigma{0.4}
\def\mua{2}
\def\mub{3}
\def\vertdislab{1.15}

\pgfmathdeclarefunction{gauss}{2}{%
	\pgfmathparse{exp(-((x-#1)^2)/(2*#2^2))}%
}


\begin{tikzpicture}
\begin{axis}[
no markers, domain=0:7, samples=200,
axis lines*=left, xlabel=$x$, ylabel={},
every axis y label/.style={at=(current axis.above origin),anchor=south},
every axis x label/.style={at=(current axis.right of origin),anchor=west},
%
%
height=6.180cm, width=10cm,
%
%
xtick = {\mua,  \mub, \mua+\mub}, 
xticklabels={
      \textcolor{my_cp_col2}{$v$},  
      \textcolor{my_cp_col4}{$u$},
      \textcolor{my_cp_col6}{$(v+u)$}
                   },
ytick=\empty,
enlargelimits=false, clip=false, axis on top,
axis lines = middle,
grid = major
]


\addplot [very thick,my_cp_col2] {gauss(\mua,\sigma)};
\path (axis cs:\mua,\vertdislab) ++ (0.0, 0.6) node [color=my_cp_col2] (a1) {$k_{v}(x)$}; 

\addplot [very thick,my_cp_col4] {gauss(\mub,\sigma)};
\path (axis cs:\mub,\vertdislab) ++ (0.0, 0.6) node [color=my_cp_col4] (a1) {$k_{u}(x)$};

\addplot [very thick,my_cp_col6] {gauss(\mua+\mub,\sigma)};
\def\muc{\mua+\mub}
\path (axis cs:(\muc,\vertdislab) ++ (0.0, 0.6) node [color=my_cp_col6] (a1) {$k_{v}(x-u)=k_{v}(x)\ast k_{u}(x)$};


\addplot+[mark=none,
        domain=\muc-\sigma:\muc+\sigma,
        samples=100,
        pattern=north east lines,
        area legend,
        pattern color=my_cp_col6]{gauss(\muc,\sigma)} \closedcycle;

\end{axis}

\end{tikzpicture}
		\caption{}
		\label{fig_rotation_image_example}
	\end{subfigure}
 \hfill
	\begin{subfigure}{.48\textwidth}
		\centering

\definecolor{my_cp_col1}{RGB}{253, 231, 37}
\definecolor{my_cp_col2}{RGB}{180, 222,44}
\definecolor{my_cp_col3}{RGB}{94, 201, 98}
\definecolor{my_cp_col4}{RGB}{33, 145, 140}
\definecolor{my_cp_col5}{RGB}{59, 82, 139}
\definecolor{my_cp_col6}{RGB}{68, 1, 84}

\definecolor{my_cp2_col1}{RGB}{255,255,217}
\definecolor{my_cp2_col2}{RGB}{237,248,177}
\definecolor{my_cp2_col3}{RGB}{199,233,180}
\definecolor{my_cp2_col4}{RGB}{127,205,187}
\definecolor{my_cp2_col5}{RGB}{65,182,196}
\definecolor{my_cp2_col6}{RGB}{29,145,192}
\definecolor{my_cp2_col7}{RGB}{34,94,168}
\definecolor{my_cp2_col8}{RGB}{37,52,148}
\definecolor{my_cp2_col9}{RGB}{8,29,88}

\def\sigma{0.4}
\def\sigmab{1.4142*\sigma}
\def\mua{2}
\def\mub{3}
\def\smalldelta{0.2}

\def\vertdislab{1.15}

\pgfmathdeclarefunction{gauss}{2}{%
	\pgfmathparse{exp(-((x-#1)^2)/(2*#2^2))}%
}

\pgfmathdeclarefunction{gaussb}{2}{%
	\pgfmathparse{(\sigma*1.7724)*exp(-((x-#1)^2)/(2*#2^2))}%
}


\begin{tikzpicture}
\begin{axis}[
no markers, domain=0:7, samples=200,
axis lines*=left, xlabel=$x$, ylabel={},
every axis y label/.style={at=(current axis.above origin),anchor=south},
every axis x label/.style={at=(current axis.right of origin),anchor=west},
%
%
height=6.180cm, width=10cm,
%
%
xtick = {2,  3, 5}, 
xticklabels={
      \textcolor{my_cp_col2}{$v$},  
      \textcolor{my_cp_col4}{$u$},
      \textcolor{my_cp_col6}{$(v+u)$}
                   },
ytick=\empty,
enlargelimits=false, clip=false, axis on top,
axis lines = middle,
grid = major
]

\addplot [very thick,my_cp_col2] {gauss(\mua,\sigma)};
\path (axis cs:\mua,\vertdislab) ++ (0.0, 0.6) node [color=my_cp_col2] (a1) {$k_{v}(x)$}; 

\addplot [very thick,my_cp_col4] {gauss(\mub,\sigma)};
\path (axis cs:\mub,\vertdislab) ++ (0.0, 0.6) node [color=my_cp_col4] (a1) {$k_{u}(x)$};

\addplot [very thick,my_cp_col6] {gaussb(\mua+\mub,\sigmab)};
\def\muc{\mua+\mub}
\path (axis cs:\muc,\vertdislab) ++ (0.0, 0.6) node [color=my_cp_col6] (a1) {$(k_{v}\star k_{u})(x)$}; 

\addplot [fill=my_cp_col2,opacity=0.5, draw=none, domain=\muc-\sigmab:\muc+\sigmab] {gaussb(\muc,\sigmab)} \closedcycle;

\addplot+[mark=none,
        domain=\muc-\sigma:\muc+\sigma,
        samples=100,
        pattern=north east lines,
        area legend,
        pattern color=my_cp_col6]{gaussb(\muc,\sigmab)} \closedcycle;


\end{axis}

\end{tikzpicture}
		\caption{}
		\label{fig_asp_interpolation}
	\end{subfigure}
 %
 %
 \caption{Comparison between the RKHS convolution ``$\ast$" in~\eqref{eq_A_rkhs_prod_ij} using $k_{v}(x)=\exp{\left( -(x-v)/(2\sigma^{2})\right)}$, and the standard ``$\star$" convolution in~\eqref{eq_classical_conv}. In the figure, $v,u\in\ccalX\subset\mbR^{+}$,~$\sigma>0$ are fixed and $(v+u)\in\ccalX$.  Left: the ``$\ast$" convolution between $k_{v}(x)$ and $k_{u}(x)$ produces $k_{v}(x)\ast k_{u}(x)=k_{v+u}(x)=k_{v}(x-u)$. The shaded patterned area under $k_{v}(x-u)$ emphasizes the interval, $[(v+u)-\sigma, (v+u)+\sigma]$, that contains $68.27\%$ of $k_{v}(x-u)$'s energy. Right: The classical convolution, ``$\star$", between $k_{v}(x)$ and $k_{u}(x)$ results in a Gaussian function, $k_{v}(x)\star k_{u}(x)$, whose amplitude is
 smaller than that of $k_{v}(x)$ and whose variance is larger than the variance of $k_{v}(x)$. The shaded area in green color emphasizes the interval, $[(v+u)-\sqrt{2}\sigma, (v+u)+\sqrt{2}\sigma]$, that contains $68.27\%$ of $k_{v}\star k_{u}$'s, and the shaded patterned area highlights the interval $[(v+u)-\sigma, (v+u)+\sigma]$ as a reference.}
 \label{fig_rkhs_conv_vs_star_gauss}
\end{figure*}
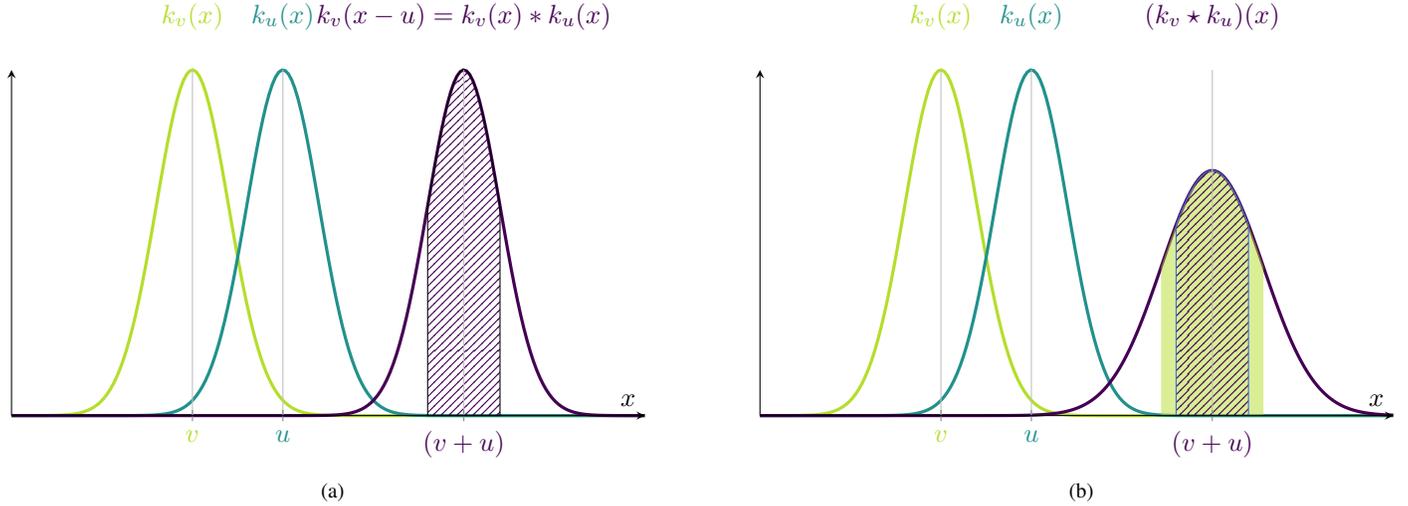




\vspace{3mm}
\subsubsection{\underline{Example -- Bandlimited Signals in Time}}
\label{ex_sincaskernel} 

Let $\ccalH$ be the space of bandlimited one-dimensional signals with bandwidth $B$~\cite[p. 10]{paulsen2016introduction}. Endowing $\ccalH$ with the $L_{2}$ inner product $\langle f, g \rangle=\int_{-\infty}^{\infty}f\overline{g}dx$, the space $\ccalH$ is an RKHS with reproducing kernel 
\begin{equation}\label{eq:sinckit}
K(u,v)
      =
       \frac{B}{\pi}
           \sinc\left(
                   \frac{B}{\pi}
                       \left( 
                          u-v
                       \right)   
                 \right)
       .
\end{equation}
To perform convolutions in $\ccalH$ we can take into account~\eqref{eq_A_rkhs_prod_ij}. Then, the convolution between a signal $f(x)=\sum_{v\in\ccalV}\alpha_{v}k_{v}(x)$ and a filter $g(x)=\sum_{u\in\ccalU}\beta_{u}k_{u}(x)$ results in a signal $h\in\ccalH$ given by
\begin{equation}\label{eq_ex_sincaskernel_2}
h(x)=
\left(g\ast f \right)(x)
     = 
      \sum_{v\in\ccalV, u\in\ccalU} 
            \alpha_{v}
            \beta_{u}
            \frac{B}{\pi}
            \sinc\left(
                       \frac{B}{\pi}
                           \left(
                              x - \left(v+u\right)
                           \right)
                  \right)
                  ,
\end{equation}
where $\ccalV,\ccalU\subset\mbR$.

The convolution in~\eqref{eq_ex_sincaskernel_2} closely relates to the standard \textit{shift} convolution, ``$\star$", given by
\begin{equation}\label{eq_classical_conv}
\left( 
       f\star g
\right)(x)       
    =
    \int_{-\infty}^{\infty}
         f(\tau)
         g(x-\tau)
         d\tau
         ,
\end{equation}
and used in classical signal processing. This is a consequence of the shift equivariance properties of ``$\star$". To see this, let us recall that
%
%
%
\begin{multline}\label{eq_star_conv_sinc}
\sinc\left(
           \frac{B}{\pi}
                \left( 
                    x-v
                \right)    
     \right)
\star
\sinc\left(
           \frac{B}{\pi}
                \left( 
                    x-u
                \right)    
     \right)
         =
         \\
         \left(
               \frac{\pi}{B}
         \right) 
              \sinc\left(
                    \frac{B}{\pi}
                         \left(
                             x-(v+u)
                         \right)    
                    \right)
               .
\end{multline}
\noindent This observation has the fundamental implication of assuring that the convolution between two functions given by ``$\ast$" in~\eqref{eq_ex_sincaskernel_2} is equivalent to the operation given by ``$\star$" when a sinc kernel induces the RKHS. Due to the importance of this fact, we stated it formally as follows.


\begin{proposition}
\label{prop_starconv_vs_rkhsconv}
Let $\ccalH$ be the RKHS of bandlimited functions in $[-B,B]$ -- as in Example~\ref{ex_sincaskernel} -- with reproducing kernel $K(u,v)$ given by~\eqref{eq:sinckit}. Let ``$\ast$" be the convolution operation defined in~\eqref{eq_A_rkhs_prod_ij} and let ``$\star$" be the classical convolution operation. Then, for any $f,g\in\ccalH\bigcap L_{1}(\mbR)$, it follows that
\begin{equation}
 f\ast g 
     =
      f\star g
      .
\end{equation}
\end{proposition}

\begin{proof}
    See Appendix~\ref{sec_proof_thm_Eucld_conv}.
\end{proof}


Example~\ref{ex_sincaskernel} and Proposition~\ref{prop_starconv_vs_rkhsconv} highlight a scenario where the classical convolution and the algebraic RKHS convolution introduced in~\eqref{eq_A_rkhs_prod_ij} are identical. Beyond this identity, the fundamental insight is that the algebraic RKHS convolutions are \textit{structurally} equivalent to the classical convolutions in the sense that they can be used to leverage \textit{symmetries and equivariance properties} of a given domain, while at the same time adding other attributes such as spatial localization or limited bandwidth, which can be beneficial for certain applications.

Now, we show a glimpse of the generalization capabilities of the convolution introduced in~\eqref{eq_A_rkhs_prod_ij}. To this end, we consider the cyclic sum, $\oplus$, on a connected, compact, and bounded set $\ccalX\subset\mbR^{+}$, given by
\begin{equation}\label{eq_cyclic_sum}
v
\oplus
u
  =
   \begin{cases}
			v+u-\sup\ccalX, & \text{if}~ \left(v+u\right)\notin\ccalX\\
                v+u, & \text{if}~ \left(v+u\right)\in\ccalX
                .
   \end{cases}
\end{equation}
Then, using~\eqref{eq_cyclic_sum} we extend the notion of convolution introduced in~\eqref{eq_A_rkhs_prod_ij} as follows,
\begin{equation}\label{eq_A_rkhs_prod_ij_c}
\left(
     \sum_{v\in\ccalV}\alpha_{v}k_{v}(x)
\right)
      \ast
\left(
     \sum_{u\in\ccalU}\beta_{u}k_{u}(x)
\right)
      =
      \sum_{v\in\ccalV,u\in\ccalU}\alpha_{v}\beta_{u} k_{v\oplus u}(x)
      .
\end{equation}
%
%
%
Like in Example~\ref{ex_sincaskernel}, the operator in~\eqref{eq_A_rkhs_prod_ij_c} is a convolution in the context of algebraic signal processing (ASP) -- we show this in Section~\ref{subsec_rkhs_algebras} --. This is,~\eqref{eq_A_rkhs_prod_ij_c} is a particular instantiation of a general algebraic signal model (ASM) that determines a subalgebra in the space of endomorphisms $\text{End}(\ccalH)$ of $\ccalH$. Now, we introduce an example where we use~\eqref{eq_A_rkhs_prod_ij_c} and Gaussian reproducing kernels, which provide certain advantages over the convolutional models with $\sinc$ kernels.


\begin{figure*}
      %
      %
      \centering
             \begin{subfigure}{.3\linewidth}
               \centering
                \includegraphics[width=1\textwidth]{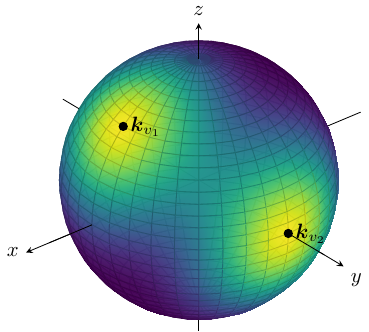}
             \end{subfigure}
                \begin{subfigure}{.3\linewidth}
                  \centering
                  \includegraphics[width=1\textwidth]{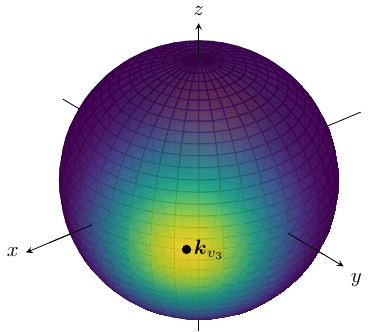}
      \end{subfigure}
                \begin{subfigure}{.3\linewidth}
                  \centering
        \includegraphics[width=1\textwidth]{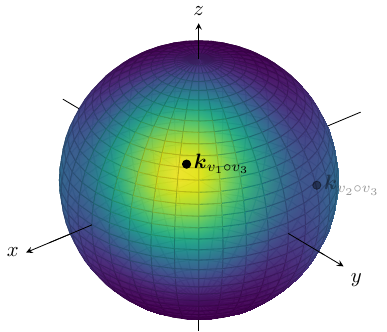}
      \end{subfigure}
      \caption{
      Depiction of the $k_{v}(x)$ functions of the RKHS on the sphere $\mbS^{2}$ as per Example~\ref{ex_rkhs_on_sphere_convfilt}. For any $u,v\in \mbS^{2}$ we have $k_{v}(u)=\left\langle u, v\right\rangle^{4}$, where $\langle\cdot,\cdot \rangle$ is the ordinary inner product in $\mbR^{N}$. Left: we show $f(u)= k_{v_1}(u) + k_{v_2}(u)$ with $v_1 =[1,0,1](1/\sqrt{2})$ and $v_2 = [0,1,0]$. Center: we show $g(u)=k_{v_3}(u)$ with $v_3 =[1,1,0](1/\sqrt{2})$. Right: we depict $f\ast g=k_{v_{1}\circ v_{3}}+k_{v_{2}\circ v_{3}}$, where $\circ$ is the group product in $\mathsf{SO}(3)$. The points $v_1$ and $v_2$ are rotated according to $v_3$, which constitutes a rotation of $45^{\circ}$ around the $z$ axis.
      }
      \label{fig_rkhs_on_sphere}
\end{figure*}




\vspace{3mm}
\subsubsection{\underline{Example -- RKHS Convolutions with Gaussian Kernels}}
\label{ex_gaussian_kernel}

Let $\ccalX\subset\mbR^{+}$ be connected, bounded and compact, and let $\ccalH$ be the RKHS of functions on $\ccalX$ induced by the Gaussian reproducing kernel
\begin{equation}\label{eq:gaussiankernel}
    K(u,v) 
         = 
           \exp \left( 
                     -B(u-v)^2
                \right)
                .
\end{equation}
In the same vein that $B$ relates to smoothness for signals expanded by $\sinc$ functions in Example~\ref{ex_sincaskernel}, the smoothness of the function represented in terms of~\eqref{eq:gaussiankernel} is represented by the parameter $B$. Low values of $B$ result in wider Gaussian bells in the expansion, and high values of $B$ give rise to fast-changing functions -- see Fig.~\ref{fig_rkhs_conv_vs_star_gauss} --. The kernel~\eqref{eq:gaussiankernel} provides advantages over~\eqref{eq:sinckit} since Gaussian kernels are more localized and do not alternate sign. Using~\eqref{eq:gaussiankernel} we can perform convolutions following~\eqref{eq_A_rkhs_prod_ij_c} while leveraging translation symmetries. To see this, notice that if $(v+u)\in\ccalX$ it follows that
\begin{equation}
k_{v}(x) 
    \ast
k_{u}(x)
     =
     k_{v\oplus u}(x)
     =
     k_{v}(x-u)
     .
\end{equation}
Then, filtering $f(x)=\sum_{v\in\ccalV}\alpha_{v}k_{v}(x)$ with $g(x)=\sum_{u\in\ccalU}\beta_{u}k_{u}(x)$ leads to the signal $h(x)$ given by
\begin{equation}
h(x) =
      \sum_{v\in\ccalV,u\in\ccalU}
               \alpha_{v}
               \beta_{u}
                   k_{v}(x-u)
      =
        \sum_{u\in\ccalU}
             \beta_{u}
             f(x-u)
             ,
\end{equation}
when $(v+u)\in\ccalX$ for all $v\in\ccalV, u\in\ccalU$. The resulting signal $h(x)$ highlights essential characteristics of convolutions based on Gaussian Kernels. It shows that the resolution of the filtered signal is determined as a superposition of functions of localized behavior whose support and amplitude are not modified by the convolution itself -- see Fig.~\ref{fig_rkhs_conv_vs_star_gauss} (Left) --. This contrasts with~\eqref{eq_classical_conv}, where the result of making a convolution of two Gaussians with the same variance and amplitude leads to a Gaussian with a larger variance and reduced amplitude -- see Fig.~\ref{fig_rkhs_conv_vs_star_gauss} (Right). Therefore, the RKHS convolutional model in Fig.~\ref{fig_rotation_image_example} facilitates the use of sparsity properties of the signals under consideration when expanded in terms of the $k_{v}(x)$. To see this, notice that those signals with a \textit{small and highly disconnected} support can be conveniently represented as a linear combination of $\{ k_{v}(x) \}_{v}$, where the $k_{v}(x)$ have a \textit{small connected} support. Such representation can be considered sparse when compared to the number of functions $k_{v}(x)$ needed to represent a function whose support equals the whole domain $\ccalX$. Convolutions based on Gaussian Kernels are also analog to the classical convolution ``$\star$" in the sense that $h(x)$ is a weighted sum of shifted versions of $f(x)$.

Notice that RKHS convolutions based on Gaussian kernels extend to multiple dimensions, i.e. $\mbR^n$, by substituting  $(v-t)^T\mathbf B(v-t)$   for $B(v-t)^2$ in  \eqref{eq:gaussiankernel} with weighting matrix $\mathbf B$. In particular, we will use $n=2$ for the experiments in section \ref{sec:num_exp}.


\section{Generalized Convolutions in RKHS}
\label{sec_nonEuclid_conv_rkhs}

This section discusses the generalized notion of convolution in RKHS spaces. We show that the RKHS convolutional models discussed in Section~\ref{sec_conv_filt_rkhs}, are  particular cases of a general algebraic signal model (ASM) that encapsulates the properties of the RKHS as a Hilbert space, and the algebraic properties of the domain. From now on, and for reasons that will become clear in Subsection~\ref{subsec_rkhs_algebras}, we will refer to the functions in any RKHS as \textit{signals}.

We start introducing the definition of the general algebraic convolution product in an arbitrary RKHS, which is a generalization of~\eqref{eq_A_rkhs_prod_ij} and~\eqref{eq_A_rkhs_prod_ij_c}.


\begin{definition}\label{def_gen_RKHS_conv}

Let $\ccalH$ be an RKHS with reproducing kernel $K(u,v)$ and let $\ccalX$ be the domain of the signals in $\ccalH$. Let $\circ: \ccalX \times \ccalX \to \ccalX$ be a binary operation in $\ccalX$ and  $\delta\in\ccalX$ such that
 \begin{equation}\label{eq:associativity}
      \left( 
            v\circ u 
      \right)\circ \ell 
                   = 
                      v\circ \left(
                                     u\circ \ell
                               \right)
                                        \quad \forall~v,u,\ell\in\ccalX,
\end{equation}
 and 
 \begin{equation}\label{eq:neutral_element}
          \delta\circ v
                 = 
                   v \circ \delta
                 = 
                   v\quad \forall~v\in\ccalX
       .
 \end{equation}
Then, the RKHS convolution product ``$\ast$" in $\ccalH$ is given by
\begin{equation}\label{eq:A_rkhs_gen_prod}
\left(
       \sum_{v\in\ccalV}\alpha_{v}k_{v}(x)
\right)
        \ast
\left(
       \sum_{u\in\ccalU}\beta_{u}k_{u}(x)
\right)
         =
            \sum_{v\in\ccalV,u\in\ccalU}\alpha_{v}\beta_{u} k_{v\circ u}(x)
            ,
\end{equation}
where $\ccalV,\ccalU\subset\ccalX$.
\end{definition}


\noindent Notice that the binary operation ``$\circ$" is associative, and $\delta$ is an identity element under ``$\circ$". From~\eqref{eq:A_rkhs_gen_prod} we can see that by choosing ``$\circ$" as the ordinary addition operation we obtain~\eqref{eq_A_rkhs_prod_ij}. This follows from the fact that ``$\circ$" endowed with~\eqref{eq:associativity} and~\eqref{eq:neutral_element} encapsulate the properties of a \textit{monoid}~\cite{awodey2010category,fong2019invitation,spivak2014category,riehl2017category}, which is an algebraic object that generalizes the notion of groups and semigroups. The well-known sets $\mbR$, $\mbN$ and $\mbZ$ with the usual addition operation and ``$0$" as the identity element, are particular cases of a monoid. Then, utilizing ``$\circ$" we attach structural properties of $\ccalX$ to the convolutional product ``$\ast$" in~\eqref{eq:A_rkhs_gen_prod} and the reproducing property of the RKHS provides the means to describe the information defined \textit{on} $\ccalX$.

Now, we present several examples of RKHS convolutional models on arbitrary domains that can be obtained as particular instantiations of Definition~\ref{def_gen_RKHS_conv}.



\vspace{3mm}
\subsubsection{\underline{Example -- Component-wise scalings in $\mbR^2$}}
\label{ex_comp_scale_in_R2}

Let $\ccalX\subset\mbR^{2}_{+}$ be connected, compact, and bounded and let $\ccalH$ be the RKHS of signals on $\ccalX$ with a Gaussian reproducing kernel. Then, we obtain a convolution product as in~\eqref{eq:A_rkhs_gen_prod} by selecting $\circ$ as the component-wise scalar product between the centers of the $k_{v}(x)$ functions. This is, if $v,u\in \mathbb R^2$ with $v=\left( v_x, v_y \right)$ and $u=\left( u_x, u_y \right)$, then we have that
\begin{equation}
\left( v_x, v_y \right)
    \circ
\left( u_x, u_y \right) 
             = 
              \left(
                   v_{x}u_{x}
                   ,
                   v_{y}u_{y}
              \right)
              ,
\end{equation}
which determines a convolution product between $\sum_{v\in\ccalV}\alpha_{v}k_{v}(t)$ and $\sum_{u\in\ccalU}\beta_{u}k_{u}(t)$ given by
\begin{equation}
\left(
       \sum_{v\in\ccalV}\alpha_{v}k_{v}(t)
\right)
        \ast
\left(
       \sum_{u\in\ccalU}\beta_{u}k_{u}(t)
\right)
         =
            \sum_{v,u}\alpha_{v}\beta_{u} k_{                   \left( v_{x}u_{x}
                   ,
                   v_{y}u_{y}
            \right)}(t)
            .
\end{equation}
Notice that the convolutional model discussed is a natural generalization of Example~\ref{ex_gaussian_kernel}, and therefore it allows to exploit the sparsity of signals on $\mbR^{2}$.


    \begin{figure*}
      \centering
             \begin{subfigure}{.32\linewidth}
               \centering
                \includegraphics[width=0.8\textwidth]{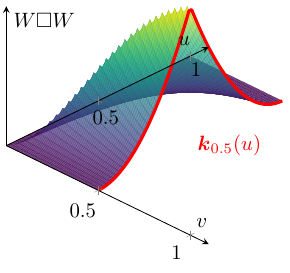}
             \end{subfigure}
                \begin{subfigure}{.32\linewidth}
                  \centering
                  \includegraphics[width=1\textwidth]{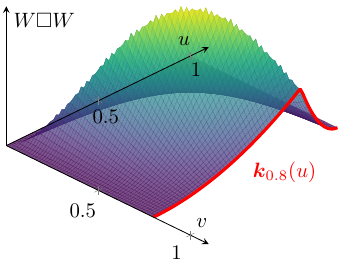}
      \end{subfigure}
                \begin{subfigure}{.32\linewidth}
                  \centering
        \includegraphics[width=0.8\textwidth]{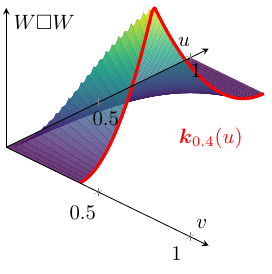}
      \end{subfigure}
      \caption{Representation of the functions $k_{v}(u)=K(u,v)$ in the RKHS graphon model considering the graphon $W(u,v)=\min(u,v)(1-\max(u,v))$. We show $K(u,v)=W\square W$ with $u$ ranging from $0$ to $1$ and $v$ ranging from $0$ to $0.5$, $0.8$, and $0.4$, respectively. We restrict the values of $v$ to emphasize the behavior of $k_{v}(u)$ concerning $K(u,v)$. We depict $k_{0.5}(u)$ in the left, $k_{0.8}(u)$ at the center, and the result of their convolution, $k_{0.5}(u)\ast k_{0.8}(u) =k_{0.5\circ 0.8} =k_{0.5\times 0.8}(u) =k_{0.4}(u)$ on the right. Notice that $v\circ u = i\times j$.}
      \label{fig_rkhs_on_graphon_ast}
      \end{figure*}




\vspace{3mm}
\subsubsection{\underline{Example -- RKHS Convolutions on the Sphere $\mbS^{N}$}}
\label{ex_rkhs_on_sphere_convfilt}

Let $\mbS^{N}\subset\mbR^{N+1}$ be the $N$-dimensional sphere embedded in $\mbR^{N+1}$, and let $\ccalH$ be the space of homogeneous polynomials of degree $d$ on $\mbS^{N}$. If $x_0,x_1, \ldots, x_N$ denote $N+1$ independent variables in $\mbR^{N+1}$, then any element $f\in\ccalH$ can be written as
$
f
   =
     \sum_{\vert\alpha\vert = d}
          w_{\alpha}
          x^{\alpha}
          ,
$
where $\alpha = \left( \alpha_0, \alpha_1, \ldots, \alpha_n\right)\in\mbN^{N+1}$ and 
$
x^{\alpha}
= x_{0}^{\alpha_0}x_{1}^{\alpha_1}\ldots x_{n}^{\alpha_n}
.
$
As shown in~\cite{cucker2002mathematical}, $\ccalH$ can be endowed with an inner product, $\langle, \rangle_{\ccalH}$ which we describe as follows. Let
$
f
    =
     \sum_{\vert\alpha\vert = d}
          w_{\alpha}
          x^{\alpha}
$
and
$
g
    =
     \sum_{\vert\alpha\vert = d}
          v_{\alpha}
          x^{\alpha}
          ,
$
then
\begin{equation}\label{eq_inner_prod_Hd}
\left\langle 
     f
     ,
    g
\right\rangle_{\ccalH}
     =
     \sum_{\vert \alpha\vert =d}
           w_{\alpha}
           v_{\alpha}
               \left( C_{\alpha}^{d}\right)^{-1}
               ,
\end{equation}
where $C_{\alpha}^{d}=d!/(\alpha_{0}!\ldots \alpha_{N}!)$. The reproducing kernel, $K(u,v)$, associated with $\ccalH$ is given by
$
 K(u,v) 
       =
        \langle 
               u, v
        \rangle^{d}   
        ,
$
for $u,v\in\mbS^{N}$ and where $\langle , \rangle$ is the ordinary inner product in $\mbR^{N+1}$. We can define in $\ccalH$ a convolutional product as in~\eqref{eq:A_rkhs_gen_prod} when considering ``$\circ$" as the group product in $\mathsf{SO}(N+1)$ -- the group of orthogonal matrices of size $(N+1)\times (N+1)$ --. This is a consequence of $\mathsf{SO}(N+1)$ being isomorphic to $\mbS^{N}$, which guarantees that for every point in $\mbS^{N}$ there is exactly one group element in $\mathsf{SO}(N+1)$. Additionally, since $\mathsf{SO}(N+1)$ is a Lie group of matrices -- rotation matrices in $\mbR^{N+1}$ -- the operation ``$\circ$" can be carried out as an ordinary matrix product. In Fig.~\ref{fig_rkhs_on_sphere} we depict an example of RKHS convolutions on $\mbS^{2}$. Notice that Example~\ref{ex_rkhs_on_sphere_convfilt} emphasizes an alternative way of performing convolutions on groups without requiring any discretization of the domain, which is the classical approach in~\cite{bruna_groupinvrepconvnn,cohen2016group,Weiler2019GeneralES,Weiler20183DSC,Worrall2017HarmonicND}.



\vspace{3mm}
\subsubsection{\underline{Example -- RKHS Convolutions on Graphons}}
\label{ex_rkhs_on_graphons}

A graphon is a symmetric bounded measurable function, $W: [0,1]^2 \to [0,1]$, that can be conceived as the limit object of a sequence of graphs of increasing size~\cite{lovaz2012large, graphon_geert}. The graphon plays the role of a limit adjacency matrix. The signals on a graphon $W(u,v)$ are identified with functions in $L_{2}([0,1])$~\cite{gphon_pooling_c,gphon_pooling_j,graphon_sampling_j}. The RKHS of signals on $W$ is determined by a reproducing Kernel, $K(u,v)$, that can be obtained from $W$ using the so-called \textit{box product}. We formalize this statement as follows.


\begin{proposition}\label{prop_WsqW}
Let $W$ be a graphon. Then, there is an RKHS of signals on $W$, $\ccalH\subset L_{2}([0,1])$,  with reproducing kernel $K(u,v)$ given by
\begin{equation}\label{eq_K_from_W}
K(u,v) =
         \left( 
              W\square W
         \right)(u,v)
       =
         \int_{0}^{1}
                W(u,z)
                W(z,v)
                    dz
             .
\end{equation}
\end{proposition}

\begin{proof}
    See Appendix~\ref{proof_prop_WsqW}.
\end{proof}


\noindent Then, using~\eqref{eq_K_from_W} we can define a convolution operation as in~\eqref{eq:A_rkhs_gen_prod} considering multiple choices of ``$\circ$". For instance, we can choose  $v\circ u := vu$ for all $v,u\in (0,1]$. This leads to the notion of convolution depicted in Fig.~\ref{fig_rkhs_on_graphon_ast}. Or we can
choose  ``$\circ$" to be the sum modulus $[0,1]$, i.e. $v \circ u = v\oplus u$ -- see~\eqref{eq_cyclic_sum}. Notice that~\eqref{eq_K_from_W} is an analog of
a matrix-matrix product in the continuum when the matrices are represented by a graphon. Additionally, it is important to emphasize that the convolutional model in Example~\ref{ex_rkhs_on_graphons} allows us to process information on a graphon while exploiting the group/monoid structure of $[0,1]$, which is not possible with the classical modeles based on polynomial diffusion~\cite{gphon_pooling_c,gphon_pooling_j,graphon_sampling_j}.


\subsection{RKHS Convolutional Algebraic Signal Models}
\label{subsec_rkhs_algebras}

This section shows that ``$\ast$" in~\eqref{eq:A_rkhs_gen_prod} from Definition~\ref{def_gen_RKHS_conv} is a formal convolution. To this end, we leverage \textit{algebraic signal processing} and we prove that~\eqref{eq:A_rkhs_gen_prod} is the product operation of a unital algebra that determines an algebraic signal model (ASM), which we formally define as follows.

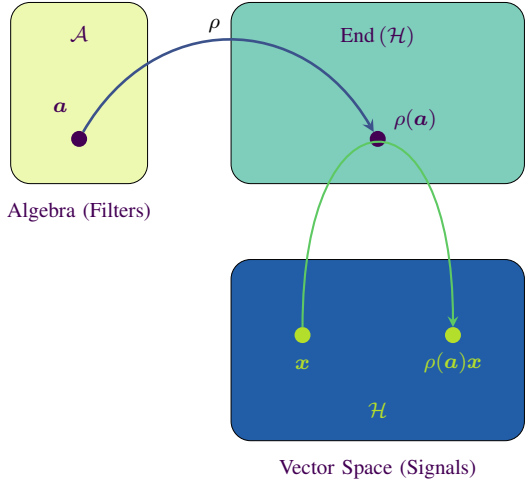
\begin{figure}
      \centering


\definecolor{my_cp_col1}{RGB}{253, 231, 37}
\definecolor{my_cp_col2}{RGB}{180, 222,44}
\definecolor{my_cp_col3}{RGB}{94, 201, 98}
\definecolor{my_cp_col4}{RGB}{33, 145, 140}
\definecolor{my_cp_col5}{RGB}{59, 82, 139}
\definecolor{my_cp_col6}{RGB}{68, 1, 84}

\definecolor{my_cp2_col1}{RGB}{255,255,217}
\definecolor{my_cp2_col2}{RGB}{237,248,177}
\definecolor{my_cp2_col3}{RGB}{199,233,180}
\definecolor{my_cp2_col4}{RGB}{127,205,187}
\definecolor{my_cp2_col5}{RGB}{65,182,196}
\definecolor{my_cp2_col6}{RGB}{29,145,192}
\definecolor{my_cp2_col7}{RGB}{34,94,168}
\definecolor{my_cp2_col8}{RGB}{37,52,148}
\definecolor{my_cp2_col9}{RGB}{8,29,88}

\usetikzlibrary{positioning,decorations.pathreplacing,shapes}


\def \scale {1.3}
\def \unit{\scale cm}
\def \layerinterdist {4}

\tikzstyle{set} = [rectangle,color=black,
                    rounded corners = 0.2*\unit,
                    fill=black,
                    inner sep=0pt,
                    draw,
                    anchor = center,
                    line width=0.1mm]
                                      
\tikzstyle{vectorspace} = [set, 
                             fill=my_cp2_col7,
                             minimum width  = 3*\unit,
                             minimum height = 1.8541*\unit]
                             
\tikzstyle{endomorphisms} = [vectorspace,
                              fill=my_cp2_col4]
                                
\tikzstyle{algebra} = [endomorphisms,
                        fill=my_cp2_col2,
                         minimum width  = 1.4*\unit,
                        minimum height = 1.8541*\unit]


\tikzstyle{dot} = [circle,
                    minimum width  = 0.12*\unit,
                    fill=black,
                    color=black,
                    inner sep=0pt,
                    draw,
                    anchor = center ]

{\fontsize{8}{8}\selectfont

\begin{tikzpicture}[rounded corners,ultra thick]


   \path (0,0) node [vectorspace, fill opacity=1] (M0) {};
   \path (M0.south) ++ (0, 0.2) node [above, color=my_cp_col2] {$\mathcal{H}$};
   
   \path (M0) ++ (-1,0.2) node [dot,color=my_cp_col2] (x) {};
   \path (x.south)++(0,-0.1) node [below, color=my_cp_col2] {$\boldsymbol{x}$}; 
   
   \path (M0) ++ (1,0.2) node [dot,color=my_cp_col2] (ex) {};
   \path (ex.south) node [below, color=my_cp_col2] {$\rho(\boldsymbol{a})\boldsymbol{x}$};

 \path (M0.south) ++ (0,0) node [] (Htext) {};
 \path (Htext.south) node [below, color=my_cp_col6,align=center] {Vector Space
   (Signals)};


   \path (M0.north) ++ (0, 1) 
         node [endomorphisms, anchor=south, fill opacity=1] (End0) {};
   \path (End0.north) ++ (0.0, -0.2) node [below, color=my_cp_col6] {$\text{End}\left(\mathcal{H}\right)$};

   \path (End0.south) ++ (0.0, 0.6) node [dot,color=my_cp_col6] (e) {};      
   \path (e.center) ++ (0.5,0.25) node [color=my_cp_col6] {$\rho(\boldsymbol{a})$}; 
   
   \path (e)+(-1,0.77) coordinate (c1);
   \path (e)+(1,0.77) coordinate (c2);   
   \path [draw, -stealth, line width=0.8,color=my_cp_col3] (x) .. controls (c1) and (c2) .. (ex);


   \path (End0.north west) ++ (-1.1, 0) 
         node [algebra, anchor = north east, fill opacity=1] (A) {};   
   \path (A.north) ++ (0,-0.2) node [below,color=my_cp_col6] {$\mathcal{A}$};
   \path (A.center) ++ (-0.25,0) node [below,color=my_cp_col6] {$\boldsymbol{a}$};  
    
    
    \path (A.south) ++ (0.0, 0.6) node [dot,color=my_cp_col6] (a1) {}; 
        
   \path (A.north) + (1,-0.1) coordinate (c1);
   \path (End0.north) + (-1,-0.1) coordinate (c2);   
   \path [draw, -stealth,line width = 1,color=my_cp_col5,  opacity=1] (a1) .. controls (c1) and (c2) .. (e) node[midway,above left,rotate=0,color=black,opacity=1] {$\rho$};

 \path (A.south) ++ (0,0) node [] (Atext) {};
 \path (Atext.south) node [below, color=my_cp_col6,align=center] {Algebra
   (Filters)};

%
%
%
%
%

\end{tikzpicture}
}
      \caption{Depiction of a generic algebraic signal model (ASM) $(\ccalA, \ccalH, \rho)$. The filters are elements of the algebra $\ccalA$, while the signals are the elements of the vector space $\ccalH$. The homomorphism, $\rho$, translates the abstract filters in $\ccalA$ into concrete linear operators in $\text{End}(\ccalH)$ that act on the signals in $\ccalH$. The symbol $\text{End}(\ccalH)$ indicates the space of linear operators from $\ccalH$ onto itself.}
      \label{fig_asp_model}
\end{figure}

\begin{definition}[\cite{parada_algnn,parada_algnnconf,algnn_nc_j,algSP0,algSP1}]\label{def_ASP}

An algebraic signal model (ASM) is defined by the triplet $(\ccalA, \ccalH,\rho)$, where $\ccalA$ is a unital algebra, $\ccalH$ is a vector space, and $\rho:\ccalA \to \text{End}(\ccalH)$ is a homomorphism, where $\text{End}(\ccalH)$ is the set of linear operators from $\ccalH$ onto itself -- see Fig.~\ref{fig_asp_model} --. 

\end{definition}


\noindent We recall that a unital algebra is a vector space with a notion of product that posses a unit or identity element with respect to such product. The filters are the elements in $\ccalA$ while the signals are the elements of $\ccalH$. The homomorphism $\rho$ maps the abstract filters in $\ccalA$ into concrete operators that act directly on the information in $\ccalH$. We recall that an algebra is a vector space that is also closed under a notion of product. Such algebra is unital when there exists a unit or identity element under the prodcut operation. Two classical examples of algebras are the set of polynomials with one independent variable and the space of square matrices. In the first case the algebra product is the standard product between polynomial expressions while in the second case is the ordinary product between matrices. The homomorphism $\rho$ in Definition~\ref{def_ASP} is nothing but a linear map that preserves the product operation between algebras. In this context, $\rho$ translates the products in $\ccalA$ into compositions of linear operators in $\text{End}(\ccalH)$.

In the light of Definition~\ref{def_ASP} we can represent a wide variety of convolutional signal models such as DTSP~\cite{algSP1}, discrete space models with symmetric operators~\cite{algSP2}, signal processing on lattices~\cite{puschel_asplattice}, signal processing on sets~\cite{puschel_aspsets}, quiver signal processing~\cite{parada_quiversp}, Lie group signal processing~\cite{lga_j,lga_icassp}, graphon signal processing~\cite{diao2016model,gphon_pooling_j,gphon_pooling_c}, multigraph signal processing~\cite{msp_j,msp_icassp2023} among others~\cite{algSP7}.

With the concept of ASM at hand, we show in the following theorem that there is a unital algebra that emerges naturally in any RKHS, and that is endowed with a product given by~\eqref{eq:A_rkhs_gen_prod}. To emphasize the domain of the functions in the RKHS $\ccalH$, we use the inclusion $\ccalH\subset\ccalF(\ccalX,\mbF)$, where $\ccalF(\ccalX,\mbF)$ is the set of functions from $\ccalX$ to the field $\mbF$ with $\mbF=\mbR$ or $\mbF=\mbC$.


\begin{theorem}\label{thm_general_A_rkhs}

Let $\ccalH\subset\ccalF(\ccalX, \mbF)$ be an RKHS with reproducing kernel $K(u,v)$. Let $\ccalS_{\ccalH}$ be the set given by
\begin{equation}\label{eq_A_rkhs_gen}
\ccalS_{\ccalH}
    =
    \text{span}
            \left(
                   \left\lbrace 
                         \left.
                               \sum_{v\in\ccalV}\alpha_{v}k_{v}(x)
                          \right\vert 
                                 k_{v}(x) = K(x,v),
                                 \ccalV\subset\ccalX ,
                                 \alpha_{v}\in\mathbb{F}
                   \right\rbrace
            \right)
    ,
\end{equation}
and let $\ast: \ccalS_{\ccalH} \times \ccalS_{\ccalH} \to \ccalS_{\ccalH}$ be the product map given by
\begin{equation}\label{eq_A_rkhs_gen_prod}
    \left( 
     \sum_{v\in\ccalV}\alpha_{v}k_{v}(x)
\right)
      \ast
\left( 
     \sum_{u\in\ccalU}\beta_{u}k_{u}(x)
\right)  
      =
      \sum_{v\in\ccalV,u\in\ccalU}\alpha_{v}\beta_{u} k_{v\circ u}(x)
      ,
\end{equation}
where $\ccalU\subset\ccalX$ and $\circ: \ccalX \times \ccalX \to \ccalX$ satisfies that 
\begin{equation}\label{eq_A_rkhs_gen_prod_associative_law}
\left(
        v \circ u
\right)
         \circ
               \ell
             =
                 v
                 \circ
                        \left( 
                                u\circ
                                \ell
                        \right)
                    .
\end{equation}
If there exists an element $\delta\in\ccalX$ such that 
\begin{equation}\label{eq_A_rkhs_gen_prod_unit_element}
\delta\circ u 
          = 
             u\circ \delta 
                       = u\quad \forall~u\in\ccalX  
 ,
\end{equation}
then the set $\ccalS_{\ccalH}$ endowed with ``$\ast$" as in~\eqref{eq_A_rkhs_gen_prod} constitutes a unital algebra that we denote by $\ccalA_{\ccalH}$.
\end{theorem}

\begin{proof}  See Appendix~\ref{proof_thm_general_A_rkhs}  \end{proof}



Theorem~\ref{thm_general_A_rkhs} provides us with the fundamental ingredient of an ASM. The elements of $\ccalA_{\ccalH}$ are the filters that will transform the signals in $\ccalH$ via the implementation carried out by $\rho$. Notice that we emphasize the span of kernel expansions $\mathcal S_{\mathcal H} \subset \ccalF(\ccalX,\mbF)$ as a different set than $\mathcal H.$ This is to address a formality in the theory of RKHSs. Although in practice, we identify $\mathcal S_{\mathcal H}$ with $\mathcal H$, our abstract definition of an RKHS makes $\mathcal S_{\mathcal H}$ only a dense subset of $\mathcal H$. Thus, we will define our algebra for signals in $\ccalS_{\mathcal H}\subset \mathcal H$ and then address this formality in Corollary~\ref{thm_density_AH_in_H} by extending it to $\mathcal H\backslash \mathcal S_{\mathcal H}$.

In the following result, we introduce the general homomorphism that implements the elements of $\ccalA_{\ccalH}$ as concrete operators in $\text{End}(\ccalH)$.


\begin{theorem}\label{thm_asm_from_rkhs}
Let $\ccalH$ be an RKHS and let $\ccalA_{\ccalH}$ be the algebra in Theorem~\ref{thm_general_A_rkhs}. Let $\rho: \ccalA_{\ccalH} \to \text{End}(\ccalH)$ be the linear map given by
\begin{equation}\label{eq_thm_asm_from_rkhs_1}
\rho\left( h \right)
            =
              h \ast (\cdot)
              ,
\end{equation}
where the action of $h \ast (\cdot)\in\text{End}(\ccalH)$ on $f\in\ccalS_{\ccalH}$ is given by $h\ast f$, wtih ``$\ast$" given as in~\eqref{eq_A_rkhs_gen_prod}. Then, the triplet $\left( \ccalA_{\ccalH}, \ccalS_{\ccalH}, \rho \right)$ is an ASM in the sense of Definition~\ref{def_ASP}.
\end{theorem}

\begin{proof} 
See Appendix~\ref{sec_proof_thm_asm_from_rkhs}. 
\end{proof}


\noindent Notice that the operation performed in~\eqref{eq_thm_asm_from_rkhs_1} on a signal $f$, determined by a representation of $h$ and $f$ as a linear combination of the functions $k_{v}(x)$, could extend uniquely to elements in $\ccalH$ that are not in $\ccalS_{\ccalH}$ under certain properties of $\ccalX$ and $\ccalA_{\ccalH}$. This is a consequence of $\ccalS_{\ccalH}$ being dense in $\ccalH$ -- see Proposition 2.1 in~\cite[p.~17]{paulsen2016introduction} --, which guarantees that for any $f(x)\in\ccalH$ and any $\epsilon>0$ there exists $\sum_{v\in\ccalV_{\epsilon}}\alpha_{v}(\epsilon)k_{v}(x)\in\ccalS_{\ccalH}$ with a countable set $\ccalV_{\epsilon}\subset\ccalX$ such that $\Vert f- \sum_{v\in\ccalV_{\epsilon}}\alpha_{v}(\epsilon)k_{v}(x) \Vert_{\ccalH}<\epsilon$. The following result formalizes this statement.


\begin{corollary}\label{thm_density_AH_in_H}
Let $(\ccalA_{\ccalH}, \ccalS_{\ccalH}, \rho)$ be the ASM in Theorem~\ref{thm_asm_from_rkhs}. Let $\epsilon>0$, $f\in\ccalH$, and $\sum_{v\in\ccalV_{\epsilon}}\alpha_{v}(\epsilon)k_{v}(x)\in\ccalS_{\ccalH}$, such that
\begin{equation}\label{eq_thm_density_AH_in_H_1}
    \left\Vert 
               f- \sum_{v\in\ccalV_{\epsilon}}\alpha_{v}(\epsilon)k_{v}(x) 
    \right\Vert_{\ccalH}
                 <
                    \epsilon
                    .
\end{equation}
If given $h\in\ccalA_{\ccalH}$ we have that $\left\Vert \rho(h) g \right\Vert_{\ccalH} \leq C_{h}\Vert g\Vert_{\ccalH}$ for all $g\in\ccalH$ with $C_{h}>0$ fixed, then it follows that
\begin{equation}\label{eq_thm_density_AH_in_H_2}
\left\Vert 
       h\ast f- h\ast\sum_{v\in\ccalV_{\epsilon}}\alpha_{v}(\epsilon)k_{v}(x) 
\right\Vert_{\ccalH}
        <
          C_{h}\epsilon
          .
\end{equation}
\end{corollary}

\begin{proof} 
See Appendix~\ref{sec_proof_thm_density_AH_in_H}.
 \end{proof}


\noindent Corollary~\ref{thm_density_AH_in_H} emphasizes in~\eqref{eq_thm_density_AH_in_H_2} that the action of $h\ast (\cdot)$ on an element $f\in\ccalH\setminus\ccalS_{\ccalH}$ is completely determined by its action on an approximation of $f$ in $\ccalS_{\ccalH}$ -- guaranteed by the density of $\ccalS_{\ccalH}$ in $\ccalH$ --. Since $\rho(h) = h\ast (\cdot)$ is a linear operator in $\text{End}(\ccalH)$, its action on any element of $\ccalH$ is well-defined. However, without Corollary~\ref{thm_density_AH_in_H}, it is not clear what is the specific effect of $\rho(h) = h\ast (\cdot)$ on $f\in\ccalH\setminus\ccalS_{\ccalH}$. The condition $\left\Vert \rho(h) g \right\Vert_{\ccalH} \leq C_{h}\Vert g\Vert_{\ccalH}$ for all $g\in\ccalH$ with $C_{h}>0$ guarantees that the images of $\ccalA_{\ccalH}$ under $\rho$ are \textit{bounded operators}.

In most practical scenarios, the ASM $(\ccalA_{\ccalH}, \ccalS_{\ccalH}, \rho)$ is enough as it is a common practice to represent the elements of $\ccalH$ by their approximation in $\ccalS_{\ccalH}$. However, if an evaluation \textit{on the limit} of the convolutional operators is necessary, one might need to consider its extension to $(\ccalA_{\ccalH}, \ccalH, \rho)$ given by Corollary~\ref{thm_density_AH_in_H}.

Notice that an RKHS of functions defined on a group naturally induces an ASM $(\ccalA_{\ccalH}, \ccalS_{\ccalH}, \rho)$ as in Theorem~\ref{thm_asm_from_rkhs}. Although this is just a particular instantiation of the many models that can be considered under Theorem~\ref{thm_asm_from_rkhs}, the fact that $\ccalX$ is a group allows us to leverage those homomorphisms $\rho$ that ensure a \textit{representation of the group} on $\ccalH$~\cite{folland2016course,hall_liealg,deitmar2014principles}, which naturally satisfy the conditions on $\rho$ in Corollary~\ref{thm_density_AH_in_H}. This is a consequence of having such representations built by definition with bounded operators~\cite{folland2016course,hall_liealg}. Information processing in groups plays an important role in physics, chemistry, signal processing, statistics, and machine learning~\cite{folland2016course,deitmar2014principles,hall_liealg,terrasFG,taylor1986noncommutative,diaconis1988group,bruna_groupinvrepconvnn,cohen2016group,Weiler2019GeneralES,Weiler20183DSC,Worrall2017HarmonicND}. This is a consequence of how groups capture the invariance of a given domain with respect to certain transformations. In physics, for instance, this is of fundamental importance as physical laws are invariant with respect to the coordinate system used to described them. In chemistry, rotation symmetries are used to characterize essential attributes in some molecules. In machine learning and signal processing, rotation and translation symmetries are used to characterize invariant geometric features of some patterns in high dimensional data. Additionally, classical signal models such as those for linear time-invariant systems, can be seen as particular case of information processing on groups where the underlying group is $\mbR$ with ordinary addition as the group product. Given the importance of information processing on groups, we formalize these ideas in the following result.


\begin{corollary}\label{cor_general_A_rkhs_group}

Let $\ccalH\subset\ccalF(\ccalX,\mbF)$ be an RKHS with reproducing kernel $K(u,v)$ and where $\ccalX$ is a group. Let $\ccalA_{\ccalH}$ be as in Theorem~\ref{thm_general_A_rkhs}, where $\circ$ is the group product and $\delta$ the group identity in $\ccalX$. Then, if $\rho: \ccalA_{\ccalH} \to \text{End}(\ccalH)$ is given by~\eqref{eq_thm_asm_from_rkhs_1}, the triplet $(\ccalA_{\ccalH}, \ccalS_{\ccalH}, \rho)$ is an ASM. Additionally, if $\rho$ also satisfies $\left\Vert \rho(h) g \right\Vert_{\ccalH} \leq C_{h}\Vert g\Vert_{\ccalH}$ for all $g\in\ccalH$ with $C_{h}>0$ fixed, then $(\ccalA_{\ccalH}, \ccalS_{\ccalH}, \rho)$ extends uniquely to the ASM $(\ccalA_{\ccalH}, \ccalH, \rho)$.

\end{corollary}

\begin{proof}  See Appendix~\ref{proof_cor_general_A_rkhs_group}  \end{proof}





\section{RKHS Neural Networks}
\label{sec_alg_rkhs_nn}

In this section, we leverage the RKHS algebraic signal models discussed in previous sections to build convolutional neural networks. To this end, we exploit the framework of algebraic neural networks (AlgNNs) introduced in~\cite{parada_algnn, parada_algnnconf, algnn_nc_j} using the ASM in Theorem~\ref{thm_asm_from_rkhs} to instantiate the convolutional operators. An AlgNN is a stacked layered structure, where information is processed in each layer employing a convolutional operator and a pointwise nonlinearity. In the following subsections, we describe the details of each of these operators.





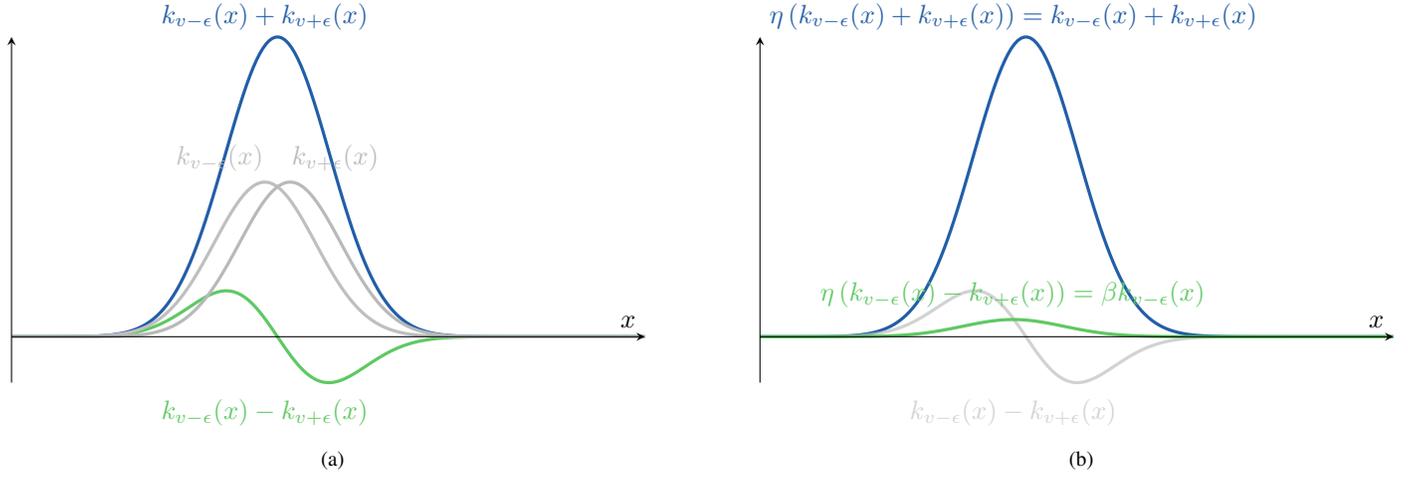
\begin{figure*}[t]
	\centering
	\begin{subfigure}{.48\textwidth}
		\centering


\definecolor{my_cp_col1}{RGB}{253, 231, 37}
\definecolor{my_cp_col2}{RGB}{180, 222,44}
\definecolor{my_cp_col3}{RGB}{94, 201, 98}
\definecolor{my_cp_col4}{RGB}{33, 145, 140}
\definecolor{my_cp_col5}{RGB}{59, 82, 139}
\definecolor{my_cp_col6}{RGB}{68, 1, 84}

\definecolor{my_cp2_col1}{RGB}{255,255,217}
\definecolor{my_cp2_col2}{RGB}{237,248,177}
\definecolor{my_cp2_col3}{RGB}{199,233,180}
\definecolor{my_cp2_col4}{RGB}{127,205,187}
\definecolor{my_cp2_col5}{RGB}{65,182,196}
\definecolor{my_cp2_col6}{RGB}{29,145,192}
\definecolor{my_cp2_col7}{RGB}{34,94,168}
\definecolor{my_cp2_col8}{RGB}{37,52,148}
\definecolor{my_cp2_col9}{RGB}{8,29,88}

\def\sigma{0.4}
\def\mua{2}
\def\mub{2.2}
\def\vertdislab{1.15}
\def\vertdislabc{0.05}

\pgfmathdeclarefunction{gauss}{2}{%
	\pgfmathparse{exp(-((x-#1)^2)/(2*#2^2))}%
}


\begin{tikzpicture}
\begin{axis}[
no markers, domain=0:5, samples=200,
axis lines*=left, xlabel=$x$, ylabel={},
every axis y label/.style={at=(current axis.above origin),anchor=south},
every axis x label/.style={at=(current axis.right of origin),anchor=west},
%
%
height=6.180cm, width=10cm,
%
%
xtick=\empty,
xticklabels={
                   },
ytick=\empty,
enlargelimits=false, clip=false, axis on top,
axis lines = middle,
grid = major
]


\addplot [very thick,my_cp2_col7] {gauss(\mua,\sigma)+gauss(\mub,\sigma)};
\path (axis cs:\mua,\vertdislab) ++ (0, 1.9cm) node [color=my_cp2_col7] (a1) {$k_{v-\epsilon}(x)+k_{v+\epsilon}(x)$};

\addplot [very thick,my_cp_col3] {gauss(\mua,\sigma)-gauss(\mub,\sigma)};
\path (axis cs:\mua,0) ++ (0, -1cm) node [color=my_cp_col3] (a1) {$k_{v-\epsilon}(x)-k_{v+\epsilon}(x)$};

\addplot [very thick,gray!50] {gauss(\mua,\sigma)};
\path (axis cs:\mua,\vertdislab) ++ (-0.6cm, 0.6) node [color=gray!50] (a1) {$k_{v-\epsilon}(x)$}; 

\addplot [very thick,gray!55] {gauss(\mub,\sigma)};
\path (axis cs:\mub,\vertdislab) ++ (0.6cm, 0.6) node [color=gray!55] (a1) {$k_{v+\epsilon}(x)$}; 

\end{axis}

\end{tikzpicture}
		\caption{}
	\end{subfigure}
 \hfill
	\begin{subfigure}{.48\textwidth}
		\centering

%
\definecolor{my_cp_col1}{RGB}{253, 231, 37}
\definecolor{my_cp_col2}{RGB}{180, 222,44}
\definecolor{my_cp_col3}{RGB}{94, 201, 98}
\definecolor{my_cp_col4}{RGB}{33, 145, 140}
\definecolor{my_cp_col5}{RGB}{59, 82, 139}
\definecolor{my_cp_col6}{RGB}{68, 1, 84}

%
\definecolor{my_cp2_col1}{RGB}{255,255,217}
\definecolor{my_cp2_col2}{RGB}{237,248,177}
\definecolor{my_cp2_col3}{RGB}{199,233,180}
\definecolor{my_cp2_col4}{RGB}{127,205,187}
\definecolor{my_cp2_col5}{RGB}{65,182,196}
\definecolor{my_cp2_col6}{RGB}{29,145,192}
\definecolor{my_cp2_col7}{RGB}{34,94,168}
\definecolor{my_cp2_col8}{RGB}{37,52,148}
\definecolor{my_cp2_col9}{RGB}{8,29,88}

\def\sigma{0.4}
\def\mua{2}
\def\mub{2.2}
\def\vertdislab{1.15}
\def\vertdislabc{0.05}
\def\myamp{0.8}

\pgfmathdeclarefunction{gauss}{2}{%
	\pgfmathparse{exp(-((x-#1)^2)/(2*#2^2))}%
}

\pgfmathdeclarefunction{gaussb}{2}{%
	\pgfmathparse{(1-\myamp)*exp(-((x-#1)^2)/(2*#2^2))/(1+\myamp)}%
}


\begin{tikzpicture}
\begin{axis}[
no markers, domain=0:5, samples=200,
axis lines*=left, xlabel=$x$, ylabel={},
every axis y label/.style={at=(current axis.above origin),anchor=south},
every axis x label/.style={at=(current axis.right of origin),anchor=west},
%
%
height=6.180cm, width=10cm,
%
%
xtick=\empty,
xticklabels={},
%
%
ytick=\empty,
enlargelimits=false, clip=false, axis on top,
axis lines = middle,
]



\addplot [very thick,my_cp2_col7,opacity=1] {gauss(\mua,\sigma)+gauss(\mub,\sigma)};
\path (axis cs:\mua,\vertdislab) ++ (0, 1.9cm) node [color=my_cp2_col7] (a1) {$\eta\left(k_{v-\epsilon}(x)+k_{v+\epsilon}(x)\right)=k_{v-\epsilon}(x)+k_{v+\epsilon}(x)$};

\addplot [very thick,gray!50,opacity=0.7] {gauss(\mua,\sigma)-gauss(\mub,\sigma)};
\path (axis cs:\mua,0) ++ (0, -1cm) node [color=gray!50,opacity=0.7] (a1) {$k_{v-\epsilon}(x)-k_{v+\epsilon}(x)$};

\addplot [very thick,my_cp_col3] {gaussb(\mua,\sigma)};
\path (axis cs:\mua,\vertdislab) ++ (-0.5, -1.8cm) node [color=my_cp_col3] (a1) {$\displaystyle\eta\left(k_{v-\epsilon}(x)-k_{v+\epsilon}(x)\right)=\beta k_{v-\epsilon}(x)$}; 

\end{axis}

\end{tikzpicture}
		\caption{}
	\end{subfigure}
 %
 %
 \caption{Effect of the pointwise nonlinearity $\eta$ in~\eqref{eq_nonlinearity_a} on the signals in $\ccalH$. For the illustration, $\ccalH$ is the RKHS of signals on $\ccalX\subset\mbR^{+}$ induced by the Gaussian reproducing kernel $K(u,v)= \exp \left(-B(u-v)^2\right)$ -- see Example~\ref{ex_gaussian_kernel} --. The signals considered are written as a linear combination of the functions $k_{v-\epsilon}(x)=K(x,v-\epsilon)$ and $k_{v+\epsilon}(x)=K(x,v+\epsilon)$ with $v,\epsilon\in\ccalX$ and where $\epsilon>0$ is small. Left: we depict the signals $g_{1}=k_{v-\epsilon}(x)+k_{v+\epsilon}(x)$ (blue) and $g_{2}=k_{v-\epsilon}(x)-k_{v+\epsilon}(x)$ (green). Right: The nonlinearity defined in~\eqref{eq_nonlinearity_a} acts on $g_{1}$ to produce $\eta\left( g_{1} \right)=k_{v-\epsilon}(x)+k_{v+\epsilon}(x)=g_{1}$ (blue). This result is a consequence of the nonnegativity of the weights in the expansion of $g_{1}$ in terms of $k_{v-\epsilon}$ and  $k_{v+\epsilon}$, and the symmetric localization of $k_{v-\epsilon}$ and $k_{v+\epsilon}$ with respect to $v\in\ccalX$. The action of $\eta$ on $g_{2}$ produces $\eta\left( k_{v-\epsilon}(x)-k_{v+\epsilon}(x)\right)=\beta k_{v-\epsilon}(x)$ (green) with $\beta=(k_{v-\epsilon}(v-\epsilon)-k_{v+\epsilon}(v-\epsilon))/(k_{v-\epsilon}(v-\epsilon)+k_{v+\epsilon}(v-\epsilon))$. Notice that $\eta$ enforces an output from $g_2$ that is nonnegative, at the expense of reducing the energy of the output signal.
 }
\label{fig_nonlinearity_on_gaussian_functions}
\end{figure*}



\subsection{The convolutional operator}
\label{sub_sec_algnn_conv_op}

The convolutional operator in the RKHS convolutional network is determined by the ASM $(\ccalA_{\ccalH}, \ccalS_{\ccalH}, \rho)$ as per Theorem~\ref{thm_asm_from_rkhs}. The data to be processed is modeled as elements of the RKHS $\ccalH$. If $h_{\ell-1}\in\ccalS_{\ccalH}$ is the input signal to the $\ell$-{th} layer of the AlgNN, the result of filtering $h_{\ell-1}$ with the convolutional operator is given by
\begin{equation}\label{eq_nn_conv_onef} 
g_{\ell}
      =
        \sum_{v\in\ccalV_{\ell}}w_{\ell,v}k_{v} \ast h_{\ell-1}
        ,
\end{equation}
where $w_{\ell,v}\mbF$ is a scalar value. In Fig.~\ref{fig:simulation_network} we depict an RKHS convolutional neural network that uses~\eqref{eq_nn_conv_onef} to perform the convolutions in the layers of the network.


\subsection{The pointwise nonlinearity}
\label{sub_sec_nonlinearity}

The point-wise nonlinearity is defined as an operator $\eta: \ccalH \to \ccalH$ whose action is \textit{point-wise} in terms of a specific basis~\cite{parada_algnn}, i.e., the action of $\eta$ is defined on the coefficients on a given basis expansion. Therefore, the images of $\eta$ are in $\ccalH$. We leverage the representation of the functions in $\ccalS_{\ccalH}$ in terms of the $k_{v}(x)$. Then, the action of $\eta$ is defined as pointwise on $g=\sum_{v\in\ccalV}\alpha_{v}k_{v}(x)$ when acting on the coefficients $\alpha_{v}$. With these notions at hand, we define the action of $\eta$ on $g\in\ccalH$ as follows
\begin{equation}\label{eq_nonlinearity_a} 
h(x)
     =
     \eta
           \left( 
                g(x)
           \right)
    =
    \sum_{v\in\ccalV}
             \frac{
       \sigma\left( 
                        g(v)
                   \right)
      }
      {
      \sum_{r\in\ccalV}k_{v}(r)
      } 
        k_{v}(x)  
        ,
\end{equation}
where $\sigma (x) = \max\{ 0, x \}$ is the traditional ReLu function. Notice that in~\eqref{eq_nonlinearity_a}, the output $h$ is a signal with the same knot functions $\{ k_{v}(x) \}_{v\in\ccalI}$ as $g(x)$. This is,~\eqref{eq_nonlinearity_a} guarantees a nonlinear transformation that is pointwise in terms of an expansion on $\{ k_{v}(x) \}_{v\in\ccalV}$.

One of the main properties of~\eqref{eq_nonlinearity_a}, is that it ensures non negativity for $g(x)$. This is illustrated in Fig.~\ref{fig_nonlinearity_on_gaussian_functions}. At the same time,~\eqref{eq_nonlinearity_a} guarantees continuity as stated in the following theorem.


\begin{theorem}\label{thm_continuity_nonlinearity}

Let $\ccalH$ be an RKHS with reproducing kernel $K(u,v)$. Then, the map $\eta: \ccalH \to \ccalH$ in~\eqref{eq_nonlinearity_a} is continuous. 

\end{theorem}
\begin{proof}
    See Appendix~\ref{sub_sec_cont_pointwisenonl}.
\end{proof}



\subsection{Training and Learnable Parameters}


\begin{figure*}
\centering
	\centering


\definecolor{my_cp5_col1}{RGB}{253, 231, 37}
\definecolor{my_cp5_col2}{RGB}{180, 222,44}
\definecolor{my_cp5_col3}{RGB}{94, 201, 98}
\definecolor{my_cp5_col4}{RGB}{33, 145, 140}
\definecolor{my_cp5_col5}{RGB}{59, 82, 139}
\definecolor{my_cp5_col6}{RGB}{68, 1, 84}

\definecolor{my_cp2_col1}{RGB}{255,255,217}
\definecolor{my_cp2_col2}{RGB}{237,248,177}
\definecolor{my_cp2_col3}{RGB}{199,233,180}
\definecolor{my_cp2_col4}{RGB}{127,205,187}
\definecolor{my_cp2_col5}{RGB}{65,182,196}
\definecolor{my_cp2_col6}{RGB}{29,145,192}
\definecolor{my_cp2_col7}{RGB}{34,94,168}
\definecolor{my_cp2_col8}{RGB}{37,52,148}
\definecolor{my_cp2_col9}{RGB}{8,29,88}


\def \myfactor {0.65}
\def \unit  {\myfactor cm}


\tikzstyle{block} = [ rectangle,
                      minimum width = \unit,
                      minimum height = \unit,
                      fill = gray,
                      draw = black,
                      text = black]

\tikzstyle{filter} = [block,
                      minimum width  = 3.5*\unit,
                      minimum height = 1.5*\unit,
                      fill = my_cp2_col7,
                      fill opacity=1,
                      text opacity=1]

\tikzstyle{nonlinearityb} = [ filter,
minimum width  = 3.5*\unit,
                             fill = my_cp2_col3,
                             fill opacity=1,
                             text opacity=1,
rounded corners = 0*\unit]

\def\deltainput     {(0,0)}
\def\deltalayer     {3}
\def\deltasigmab     {( 3.5, 0.0)}

%
\def \one   {$\textcolor{my_cp2_col2}{\displaystyle{  \boldsymbol{w}_{1}^{(1,0)}\ast f }}$}
\def \two   {$\textcolor{my_cp2_col2}{\displaystyle{\boldsymbol{w}_{2}^{(1,0)}\ast f    }}$}


{\fontsize{10}{10}\selectfont\begin{tikzpicture}[scale = \myfactor]

  \pgfdeclarelayer{bg}     
  \pgfsetlayers{bg,main}   

  \node (input) [rectangle, minimum width = 0.1*\unit] {};



  \path (input.east)  ++ \deltainput node [filter]     (L1F1) {\one};
  \path (L1F1) ++ (0,-1.3*\deltalayer) node [filter]       (L1F3) {\two};

  \path (L1F1.east) ++ \deltasigmab node [nonlinearityb] (L1Nl1)      {
  \textcolor{my_cp5_col6}{
   $
    \displaystyle h_{1}^{(1)}= \eta\left( g_{1}^{(1)}\right)
   $
   }
    };
   \path (L1F3.east) ++ \deltasigmab node [nonlinearityb] (L1Nl3)      {
   \textcolor{my_cp5_col6}{
   	$
   	\displaystyle h_{2}^{(1)} = \eta\left( g_{2}^{(1)} \right)
   	$
    }
   };



  \path (L1Nl1) ++ (2*\deltalayer,0.3*\deltalayer) node [filter]       (L2F1) {
  	\textcolor{my_cp2_col2}{
  		$
  		\boldsymbol{w}_{1}^{(2,1)}
  		\ast
  		h_{1}^{(1)}
  		$
  		}
  	};
  \path (L1Nl1) ++ (2*\deltalayer,-0.3*\deltalayer) node [filter]       (L2F3) {
  	\textcolor{my_cp2_col2}{
  		$
  		\boldsymbol{w}_{1}^{(2,2)}
  		\ast
  		h_{2}^{(1)}
  		$
  	    }
  	};


  \path (L1Nl3) ++ (2*\deltalayer,0.3*\deltalayer) node [filter]       (L2bF1) {
  	\textcolor{my_cp2_col2}{
  		$
  		\boldsymbol{w}_{2}^{(2,1)}
  		\ast
  		h_{1}^{(1)}
  		$
  	}
  	};
  \path (L1Nl3) ++ (2*\deltalayer,-0.3*\deltalayer) node [filter]       (L2bF3) {
  	  	\textcolor{my_cp2_col2}{
  	  		$
  	  		\boldsymbol{w}_{2}^{(2,2)}
  	  		\ast
  	  		h_{2}^{(1)}
  	  		$
  	  	}
  	};


  \path (L2F1) ++ (1.2*\deltalayer, -0.25*\deltalayer) node [scale=2] (oplusa){\textcolor{black}{$\oplus$}};

  \path (L2bF1) ++ (1.2*\deltalayer, -0.25*\deltalayer) node [scale=2] (oplusb){\textcolor{black}{$\oplus$}};

  \path (oplusa.east) ++ \deltasigmab node [nonlinearityb] (L2Nl1)      {
  \textcolor{my_cp5_col6}{
  	$
  	h_{1}^{(2)} = \eta\left( g_{1}^{(2)} \right)
  	$
        }
  };
  \path (oplusb.east) ++ \deltasigmab node [nonlinearityb] (L2Nl3)      {
  \textcolor{my_cp5_col6}{
  	$
  	h_{2}^{(2)} = \eta\left( g_{2}^{(2)} \right)
  	$
   }
  };

  \path (L2Nl1) ++ (\deltalayer, -0.65*\deltalayer) node [scale=2] (oplusc){\textcolor{black}{$\oplus$}};


  \path (L1F1) ++ (-1.2*\deltalayer, -0.65*\deltalayer) node [scale=1]  (x) {$f$};

  \path[draw, -stealth] (x.east) -- (L1F1.west);
  \path[draw, -stealth] (x.east) -- (L1F3.west);


  \path[draw, -stealth] (L1F1.east) --  node [above] {$g_{1}^{(1)}$} (L1Nl1.west);

  \path[draw, -stealth] (L1F3.east) --  node [above] {$g_{2}^{(1)}$} (L1Nl3.west);

  \path[draw, -stealth] (L1Nl1.east) --  node [above] {} (L2F1.west);
  \path[draw, -stealth] (L1Nl1.east) --  node [above] {} (L2bF1.west);

  \path[draw, -stealth] (L1Nl3.east) --  node [above] {} (L2F3.west);
  \path[draw, -stealth] (L1Nl3.east) --  node [above] {} (L2bF3.west);

  \path (oplusa) + (-0.28,0) coordinate (c1);
  \path[draw, -stealth] (L2F1.east) --  node [above] {} (c1);
  \path[draw, -stealth] (L2F3.east) --  node [above] {} (c1);

  \path (oplusb) + (-0.28,0) coordinate (c1);
  \path[draw, -stealth] (L2bF1.east) --  node [above] {} (c1);
  \path[draw, -stealth] (L2bF3.east) --  node [above] {} (c1);

  \path (oplusa) + (0.28,0) coordinate (c1);
  \path[draw, -stealth] (c1) --  node [above] {$g_{1}^{(2)}$} (L2Nl1.west);

  \path (oplusb) + (0.28,0) coordinate (c1);
  \path[draw, -stealth] (c1) --  node [above] {$g_{2}^{(2)}$} (L2Nl3.west);

  \path (oplusc) + (0,0.28) coordinate (c1);
  \path[draw, -stealth] (L2Nl1.east) --  node [above] {} (c1);

  \path (oplusc) + (0,-0.28) coordinate (c1);
  \path[draw, -stealth] (L2Nl3.east) --  node [above] {} (c1);


  \path (oplusc.east) ++ (1, 0) node [scale=1]  (yout) {\textcolor{black}{$f_{out}$}};

  \path (oplusc) + (0.2,0) coordinate (c1);
  \path[draw, -stealth] (c1) --  node [above] {} (yout);


  \begin{pgfonlayer}{bg}
      \path (L1F3.west |- L1F3.south) ++ (-0.4,-1.3)
           node [filter, anchor = south west,
                 fill = black!5,
                 minimum width  = 9.85*\unit,
                 minimum height = 8*\unit,]
        (layer)
        {};
        \path (L2bF3.west |- L2bF3.south) ++ (-0.4,-0.4)
           node [filter, anchor = south west,
                 fill = black!5,
                 minimum width  = 12.5*\unit,
                 minimum height = 8*\unit,]
        (layer)
        {};
    \end{pgfonlayer}

\end{tikzpicture}}
\caption{RKHS convolutional architecture used in our numerical experiments. The input signal, $f$, is processed in the first layer by $N_{1}=2$ filters whose output is later processed by the pointwise nonlinearity $\eta$. The output features from layer one are fed into a second layer, where they are first processed by two filters per input feature ($N_{2}=1$). Then, the output of such filters are combined by addition and later processed by the pointwise nonlinearity $\eta$. The outputs from the second layer are added up to obtain the output $f_{out}$.}
\label{fig:simulation_network}
\end{figure*}
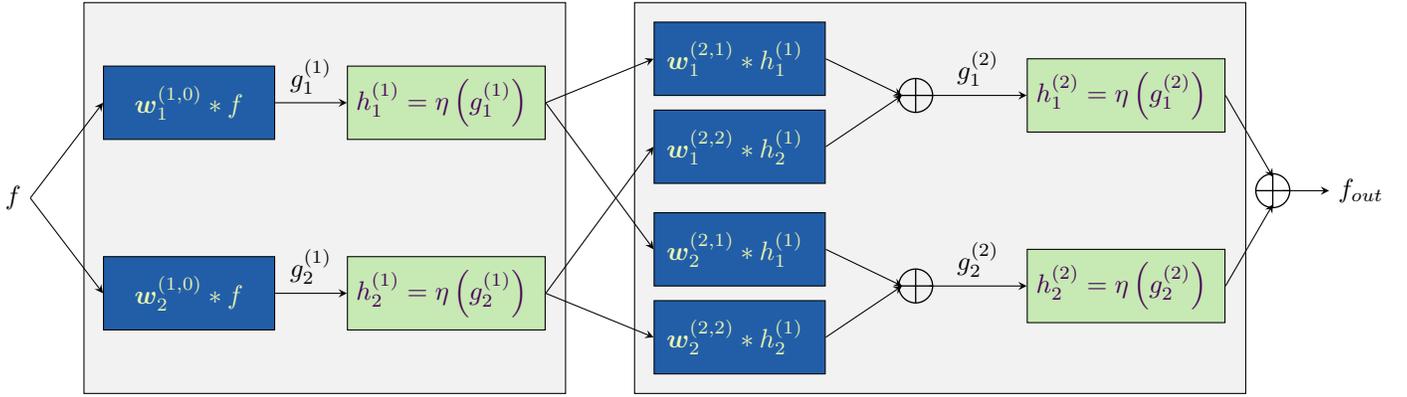


The learnable parameters in the RKHS-based network are the filters used in each layer -- which are elements of $\ccalA_{\ccalH}$ in Theorem~\ref{thm_general_A_rkhs} --. Then, our goal when training the network is to find the filters that minimize the quadratic error, measured when comparing the output of the network to the given input and reference signals. To perform our experiments and to show the full details of how the gradient descent approach can be used to find the optimal filters, we select the architecture in Fig.~\ref{fig:simulation_network} whose input-output relationship is given by
\begin{equation}\label{eq_input_output_relative_network}
f_{out}
=
\sum_{i=1}^{N_2}
  \eta\left(
         \sum_{j=1}^{N_1}
              \boldsymbol{w}_{i}^{(2,j)}
                   \ast
               \eta\left(
                       \boldsymbol{w}_{j}^{(1,0)}
                            \ast
                       f
                   \right)
   \right)
   ,
\end{equation}
where the symbol $\boldsymbol{w}_{a}^{(b,c)}$ is the $a$-th filter in the $b$-th layer processing the $c$-th feature. Fig.~\ref{fig:simulation_network} depicts a 2-layered neural network. In the first layer, we have a total of $N_1=2$ filters, $\left\lbrace \boldsymbol{w}_{j}^{(1,0)} \right\rbrace_{j=1}^{N_1}$, each followed by a point-wise non-linearity, $\eta(\cdot)$. The first layer produces $N_1=2$ features $h_{j}^{(1)}, \ j=1, \hdots, N_1$ that are fed into a second layer with $N_2=2$ filters, $\left\lbrace \boldsymbol{w}_{i}^{(2,j)} \right\rbrace_{i=1}^{N_2}$ -- $N_{2}=2$ filters per each feature coming from layer one --. We add up the output of those $N_2$ filters into the signals $g_{i}^{(2)} \ i=1\hdots N_2$ and then apply the nonlinearity $\eta(\cdot)$ to each $g_{i}^{(2)}$. The outputs of these nonlinearities are added again to obtain the final output of the network $f_{out}$.

To formulate the optimization problem that will allow us to find the optimal filters in the RKHS network, let us rewrite~\eqref{eq_input_output_relative_network} as
\begin{equation}\label{eq_input_output_relative_network_2_learn}
f_{out}
     =
      F\left(
           \boldsymbol{w}_T; f
       \right)
       ,
\end{equation}
with $\boldsymbol{w}_{T}$ collecting all the filters in the network, i.e.,
\begin{equation}
\boldsymbol{w}_{T}
    =
\left(
    \left(
        \boldsymbol{w}_{j}^{(1,0)}
    \right)_{j=1}^{j=N_1}
    ,
    \left(
        \boldsymbol{w}_{i}^{(2,j)}
    \right)_{j=1,i=1}^{j=N_1,i=N_2}
\right)
.
\end{equation}
Then, given an input $f\in\ccalH$ and a reference signal $r\in \ccalH$ we aim to minimize the quadratic error
$
\frac{1}{2}
            \left\Vert
                  r
                  -
                  f_{out}
            \right\Vert_{\calH}^2
 $, which results in the following optimization problem:
\begin{equation}\label{opt_general_rkhs_nn}
 \min_{\boldsymbol{w}_{T}}
               \frac{1}{2}
                         \left\Vert 
                                     r - f_{out}
                         \right\Vert_{\calH}^2,\quad\text{s.t.}\quad f_{out} = F\left(
                         \boldsymbol{w}_{T}; f
                      \right) .
\end{equation}
The filters $\boldsymbol{w}_{T}$ in~\eqref{opt_general_rkhs_nn} can be obtained using a gradient descent approach based on Fr\'echet derivatives and functional optimization. Such procedure is described in Appendices~\ref{sec_appendix_frechet}, and~\ref{sec_frechet_derivatives} in the supplementary material.

Since the filters can be written in terms of the functions $k_{v}(x)$ in the RKHS, learning the parameters in~\eqref{eq_input_output_relative_network} and~\eqref{opt_general_rkhs_nn} translates into learning the amplitudes and centers of each filter -- the weights multiplying the $k_{v}(x)$ functions and the values of $v$ --. This leads us to a simplified parametric approach where we restrict the filters to be determined by a finite number of parameters.

\section{Experiments and numerical simulations}
\label{sec:num_exp}

In this section, we present a numerical experiment with real data, to illustrate the use of the proposed convolutional AlgNN for learning in an RKHS. The experiment involves predicting wireless coverage on the right side of a soccer field based on measurements taken on the left side of the field. The measurements are obtained from unmanned autonomous vehicles (UAVs) that measure the wireless coverage at different locations on the field.


\begin{figure}
\centering
\input{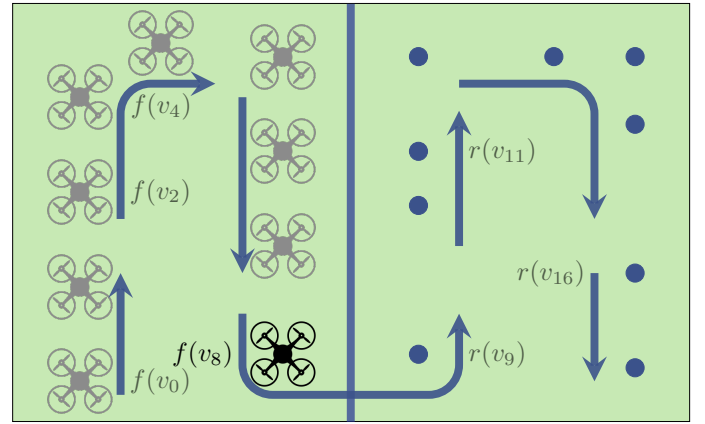}
\caption{Sketch of the simulation scenario. A UAV flies over a football field, following the path indicated by the blue arrows. A wireless coverage measure, $f$, is known on the points $\{ v_{i} \}_{i=0}^{8}$. The wireless coverage given by the signal $f(x)=\sum_{i=0}^{8}k_{v_{i}}(x)f(v_{i})$ is used to determine the wireless coverage on the arbitrary points concentrated predominantly on the right-hand side of the field. The reference signals to train the convolutional RKHS network are given by $r(x)=\sum_{i=9}^{17}k_{v_{i}}(x)r(v_{i})$, where the points $\{ v_{i} \}_{i=9}^{17}$ are arbitrarily defined on the right side of the field. Notice that  $\{ v_{i} \}_{i=0}^{8}$ and $\{ v_{i} \}_{i=9}^{17}$ are sampled on random trajectories followed by the UAV  over each side of the field .}
\label{fig_sim_scenario}
\end{figure}



\subsection{Dataset}

A dataset containing measurements of wireless coverage at different locations {for a total of 16 drone flights} is used for the numerical experiments. In each of these flights, the drone measures the wireless coverage at $18$ different locations, ranging from $-40$ to $40 \, \text{m}$ in both the horizontal and vertical coordinates, obtaining throughput measurements of up to 10Mbps.

The dataset obtained form each flight is split into two groups based on the value of the horizontal coordinate of the drone's position. The first group, {consisting of $N_S=9$ samples per flight, contains measurements where the horizontal-coordinate {is less than or equal to $0$,} and is used as the input data $f$ for the model. The second group, also consisting of $N_S=9$ samples per flight, contains measurements where the horizontal coordinate is greater than $0$,} and is used as the target data $f_{out}$ for the model. This means the objective is to predict the wireless coverage on the right side of the field based on the measurements on the left side of the field.

Of the $N_D=16$ datasets, one per flight, four of them are excluded from the training data and kept for evaluation purposes. The remaining $N_T=12$ sets are used to train the model.


\subsection{Preprocessing}

In order to work in an RKHS, the data (this is, each of the $N_D=16$ input series and each of the $N_D=16$ target series) is preprocessed to obtain a representation in the form of a Gaussian RKHS signal. That means that a representation of the form $f(x)=\sum_{v\in\mbR^2}\alpha_{v}k_{v}(x)$ is constructed from each series, where the kernel centers $v$ are chosen to be the locations of the measurements, and the coefficients $\alpha_v$ are computed from the measurement values according to the expression
\begin{equation}
\boldsymbol{\alpha} 
           = 
             \left(
                     \mathbf{K}^{\mathsf{T}}\mathbf{K}+\lambda \mathbf{K}
              \right)^{\dagger}
              \mathbf{K}
              \mathbf{f}
              .
    \label{eq:alpha}
\end{equation}
Here $\mathbf{K}$ is the kernel matrix whose entries are given according to $[\mathbf{K}]_{i,j}=K(v_{i}, v_{j})$ and  $\mathbf{f}$ is the vector of measurements with $[\mathbf{f}]_{i} = f(v_{i})$, $\lambda$ is a regularization parameter, and $(\cdot)^\dagger$ represents the Moore-Penrose pseudoinverse. The matrix $\mathbf{K}$ computed using the Gaussian kernel $K(u,v)=\exp(-\|u-v\|^2/2\sigma^2)$ for each possible pair of centers $(v_i, v_j)$, where $\sigma=10 [\text{m}]$ is the width of the kernel. The regularization parameter $\lambda$ is set to $10^{-3}$. It is straightforward to see that the expression in \eqref{eq:alpha} is the solution to the regularized least squares problem $\min_{\boldsymbol{\alpha}}\Vert \mathbf{K}\boldsymbol{\alpha}- \mathbf{f}\Vert^2+\lambda\Vert\boldsymbol{\alpha}\Vert^2$.

Notice that this preprocessing step results in a kernel expansion of the signal in terms of kernels to fit the data. In that sense, it reminds of a sinc reconstruction filter in classical signal processing. But in this case, the kernel is not a sinc, the data is assumed noisy,  and samples are not uniformly spaced. On the contrary, the wireless coverage is sampled at positions that vary from flight to flight since they are in the stochastic trajectories of the UAVs. This highlights one of the advantages of using RKHSs signals, which can naturally accommodate non-uniform sampling, since the centers in \eqref{eqn_representation_property} are not constrained to a set. This property is passed down to the AlgNN described next, which can cope with non-uniformly sampled signals like those in Figure \ref{fig_contour_plots}. This contrasts with standard neural networks, which assume that successive vector-valued inputs are intrinsically discrete or sampled from the same set.


\subsection{Model Definition and Training}

A neural network model (\texttt{AlgNN}) is defined, which consists of six two-dimensional Gaussian RKHS filters (\texttt{k101}, \texttt{k102}, \texttt{k211}, \texttt{k212}, \texttt{k221}, and \texttt{k222}). The group operator selected for this experiment is the translation or standard sum for vectors in $\mathbb R^2$. The model is trained using the Adam optimizer with a learning rate of 0.01. The training is performed over 2000 iterations, excluding four sets from the training data for evaluation purposes.

Each filter is initialized with three Gaussian kernels having an amplitude of $1$ and a width of $10$, both centered at the origin. During training, both the amplitudes and centers (but not the width) of each filter are optimized to minimize the loss function. As we will be using $l=0,\ldots,11$ pairs of input and target signals, the loss function is defined as the sum of distances between the outputs and the target signals, i.e.,
\begin{equation}
\mathrm{loss} 
            = 
               \sum_{l=0}^{11}
                        \left\Vert
                                r_l(v)-f_{l,out}(v)          \right\Vert_{\ccalH}^2, 
                        \quad
                        f_{l,out}(v) = \mathrm{AlgNN}_{\boldsymbol{w}_T}(f_l(v))
                        .
\end{equation}


\subsection{Results}


\begin{figure*}[t]
\centering
\begin{subfigure}{.16\textwidth}
    \centering
    \includegraphics[height=2.8cm]{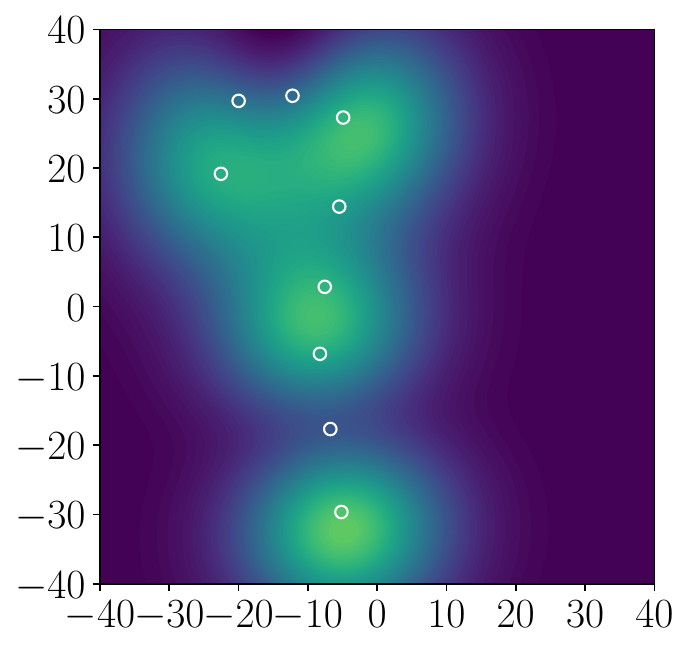}
    \caption{\footnotesize Input signal}
    \label{fig:input_7}
\end{subfigure}%
\begin{subfigure}{.16\textwidth}
    \centering
    \includegraphics[height=2.8cm]{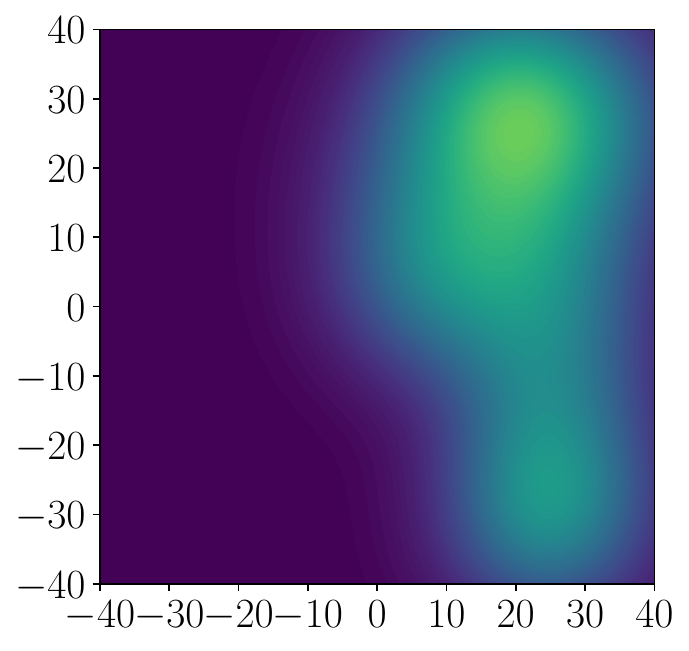}
    \caption{\footnotesize Output (AlgNN)}
\end{subfigure}%
\begin{subfigure}{.16\textwidth}
    \centering
    \includegraphics[height=2.8cm]{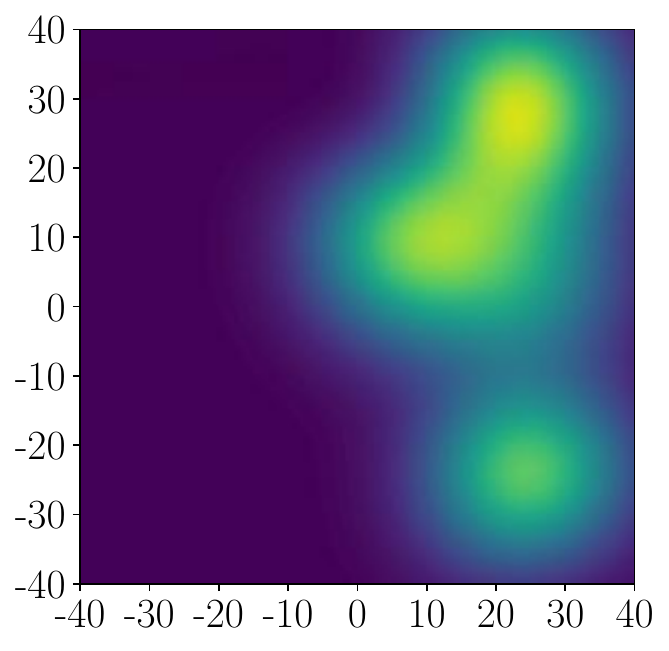}
    \caption{\footnotesize Output (KB regressor)}
    \label{fig:output_regressor}
\end{subfigure}%
\begin{subfigure}{.16\textwidth}
    \centering
    \includegraphics[height=2.8cm]{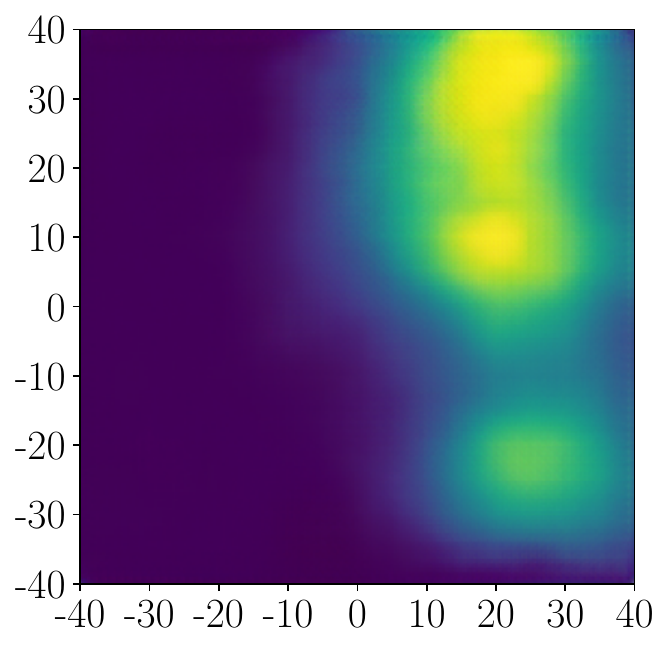}
    \caption{\footnotesize Output (pix2pix)}
\end{subfigure}%
\begin{subfigure}{.16\textwidth}
    \centering
    \includegraphics[height=2.8cm]{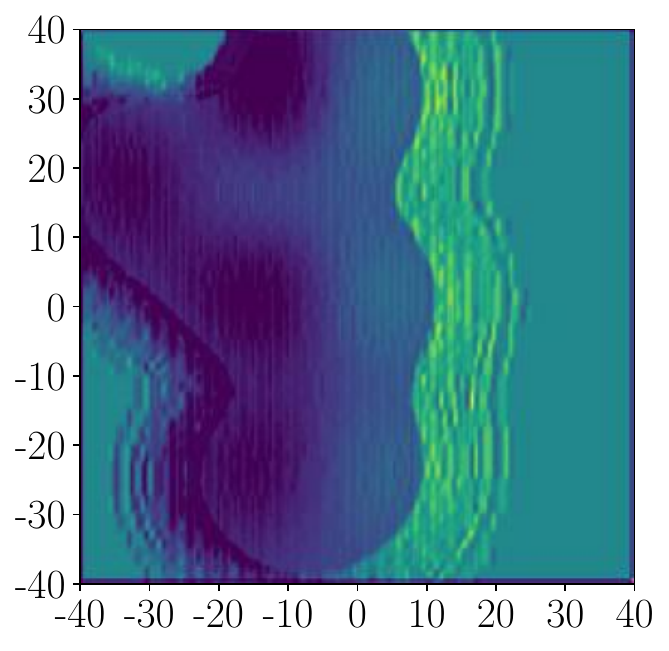}
    \caption{\footnotesize Output (2-layer CNN)}
\end{subfigure}%
\begin{subfigure}{.16\textwidth}
    \centering
    \includegraphics[height=2.8cm]{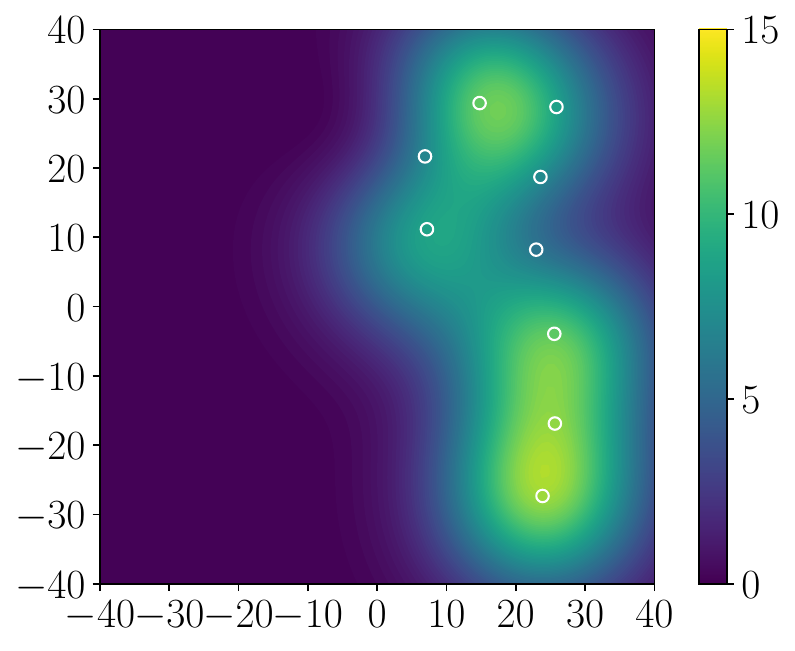}
    \caption{\footnotesize Target signal}
\end{subfigure}
\caption{Prediction of wireless coverage for a test signal. Unmanned autonomous vehicles (UAVs) measure the wireless coverage at $N=9$ waypoints on the left side of a soccer field and fit them to an RKHS signal $f\in\mathcal{H}$ (a). This signal $f\in\mathcal{H}$ is used as input for an algebraic neural network (AlgNN), whose structure is shown in Fig. \ref{fig:simulation_network}. The output $\mathrm{AlgNN}(f)\in\mathcal{H}$ (b) predicts the coverage on the right side of the field, which is the target signal shown in (f). We compare this result with the output of standard KB regressor (c) a pix2pix GAN obtaining the corresponding output in (d), and a 2-layer CNN, obtaining (e). The relative MSE averaged over the test set, results in $0.0646,0.0795,0.1059,0.2053$ for the AlgNN, KB regresssor, pix2pix, and CNN respectively.}
\label{fig_contour_plots}
\end{figure*}


{}{
Plots of the input, output, and target signals are presented in Figs. \ref{fig_contour_plots}~(a), \ref{fig_contour_plots}~(b), and \ref{fig_contour_plots}~(e), respectively. The input and target signals correspond to one of the four flights in the evaluation set. As the figure shows, the model is able to capture the underlying patterns in the data, as the output signals approximate the target signal with a relative mean square error (MSE) of $0.0645$. 
}

{}{
The performance of our AlgNN predictor is compared to that of a state-of-the-art convolutional neural network (CNN). Specifically, we trained the Conditional Adversarial Network (CAN) architecture implemented by  pix2pix  \cite{CycleGAN2017,isola2017image}. In this case, the training input-target pairs are RGB images representing the RKHS reconstructions of the data that we used for training our AlgNN. The output of the trained pix2pix for the input image in Fig. \ref{fig_contour_plots} (a) is presented in Fig \ref{fig_contour_plots} (d). The relative MSE obtained by pix2pix on the test set of $4$ images amounted to $0.1059$. The comparative lower error of our AlgNN corroborates the capacity of our RKHS filters and non-linealities to capture the information in the signal model. In contrast,  pixel-based filters and nonlinearities of pix2pix do not capture such structure but must learn signals in a space of significantly higher dimensions, which results in lower performance even with a deeper architecture.
}

{For a fairer comparison, we also trained a CNN with a simpler architecture. One comparable to that of our AlgNN in Fig. \ref{fig:simulation_network}.  As in Fig. \ref{fig:simulation_network}, we used two layers with two neurons per layer, and we aggregated the two outputs of the second layer (pixel-wise) into a single output image. Each of the six filters in this network is implemented by convolving with a $3\times3$ matrix and adding a bias level. This number of parameters per filter is comparable to the nine dimensions ($3$ centers and $3$ amplitudes) of our AlgNN filters.  The output of each image filter passes through standard ReLU nonlinearity. The input-target images for this simplified CNN are one-channel (grayscale) renditions of the signals interpolated from the training data. After training, we input the test signal in  Fig.  \ref{fig_contour_plots} (a), obtaining the image in Fig. \ref{fig_contour_plots} (e). The grayscale output is rendered using the viridis colormap for consistency with the presentation of the previous results.     Fig. \ref{fig_contour_plots} (e) shows that since the standard ($3\times3$) image kernels have only a local effect, the image convolutions cannot ``move'' the input image from the left side to the right side of the field. So, the best the CNN  can do with two layers is to add a constant bias to the whole image and then reduce its effect on the left half of the field by subtracting the local convolutions with the input.  Hence, the performance of this architecture is noticeably degraded with a relative MSE of $0.2053$.} 
{}{All MSE figures were computed by taking the Frobenius-norm square of the difference between the output and target images and dividing by the Frobenius-norm square of the target. These relative errors were averaged across the four signals in the evaluation set.  The Frobenius norm was chosen instead of the RKHS norm so that the images produced by both the pix2pix network and the simpler CNN could be evaluated. Even the figure of $0.0646$ obtained by our AlgNN was computed using the Frobenius norm after rendering the signals as RGB images. Notice that the ALgNN obtained a higher performance in terms of this image-related Frobenius MSE, even if our network was trained to minimize a different distance defined in terms of the RKHS norm.}

This second comparison to a simpler CNN further highlights the different nature of our RKHS filters when compared to the standard Euclidian convolution. Our filters take advantage of the flexibility of the group operators -- the $\circ$ operator in Definition \ref{def_gen_RKHS_conv}-- to perform different transformations on the signal domain. This theoretical innovation becomes of practical relevance as our filters are able to perform the required translations on the signals from the left to the right side of the field, thus promoting a relatively high performance of the AlgNN even with minimal architecture.

Finally, we included a comparison with a standard kernel-based regressor. In this case, model a function $f(z,x)$ where $x\in \mathbb R^2$ represents a position in the area of interest, and $z\in  \mathbb R^{N_S} \times \mathbb R^{2N_S}$ represents the information collected by the UAV on the left side of the field, including the $N_S$ measurements of wireless coverage and the $N_S$ positions where these measurements are acquired.  Thus $f(z,x)$ can predict the wireless coverage at a point $x$ given the partial information $z$ acquired during a flight, which is the same goal that we set for our AlgNN. In this new model, the predictor $f(z,x)$ belongs to an RKHS associated with a Gaussian kernel $K((z,x),(z^\prime,x^\prime))$ with $\sigma=10$, and fits the dataset of $N_S=9$ points $x$ per flight for a total of  $N_T=12$ collections $z$ obtained during the training flights. The fitting procedure results in an expansion $f(z,x)=\sum_{i=1}^{N_SN_T}\alpha_i K((z,x),(z_i,x_i))$ with coefficients given by \eqref{eq:alpha}. After training, we predict the wireless coverage for each of the $N_D-N_T=4$ datasets reserved for testing. The result for one of these test datasets is illustrated in Fig. \ref{fig_contour_plots} (c), and the relative MSE over the four test flights resulted in  $0.0795$. We claim that this sightly worse performance with respect to the AlgNN is due to overfitting. Indeed, the predictor  $f(z,x)$ aggregates $N_S N_T=108$ kernel bases, compared to a total of $18$ bases in the six filters that compound our AlgNN. This count evidences that, even if the performance is comparable, the standard kernel regression does not scale well with the data, since the number of bases grows linearly with both the number of training flights and acquisition points, as opposed to the AlgNN structure which remains unchanged as new data are processed. Training the kernel regressor presents a higher computational complexity, not only related to the number of parameters to fit, but also because computing $K((z,x),(z_i,x_i))$ involves the norm of vectors $(z,x)\in \mathbb R^{3  N_S+2}$, compared to the variables $v\in \mathbb R^2$ in the AlgNN filters.   Furthermore, the kernel regressor model is limited to the case in which the UAV collects the same number of samples $N_S$ per flight, while the AlgNN can accommodate heterogeneous dimensions because it models the input as a field of continuous domain instead of a vector.  
Once again, the filters in the AlgNN have the advantage of using group operators to achieve better performance with a minimal architecture that offers more flexibility in terms of the signals it can process.
%
%




\section{Discussion and Conclusions}
\label{sec_discussion_conclusion}

We have shown that convolutional signal models emerge naturally from an arbitrary RKHS. In particular, we proved that under the ASP framework, there is a general algebra of filters that encapsulates the algebraic (monoid or group) structure of the signal domain -- i.e., the domain of the functions in the RKHS -- while preserving the representation power of the RKHS.
The convolutional filters in the algebra operate on the elements (the signals) of an RKHS, $\ccalH$, giving, as a result, new signals that belong to $\ccalH$ as well.

The particular instantiations of the convolutional RKHS are rich and diverse as they allow for new types of convolutional operators on shift-invariant spaces, signals on groups, and graphon signals. At the same time, such generality is consistent with convolutional operators on classical signal spaces. In this line of thinking, our results and examples also showed that we can use different kernels -- and consequently different RKHS spaces -- on the same domain. This allows us to formally leverage the symmetries of a given domain while having filters with different localization attributes. This aspect is emphasized in Examples~\ref{ex_sincaskernel} and~\ref{ex_gaussian_kernel}, where the convolutional operators leverage time-shift invariances, without increasing the width of  Gaussian functions, same as sinc functions preserve bandwidth when passed through a standard convolution.

Let us highlight that the possibility of selecting a convolutional model with multiple kernel choices has direct benefits on the computational cost of designing the filters. Specifically, choosing kernels whose energy is highly concentrated on small regions in the domain of the signals ensures that one has the ability to design highly sparse filters. Additionally, tuning the kernel's parameters in the RKHS allows for a proper tradeoff between sparsity and smoothness. All these aspects open the door for future research directions on filter bank design relying on RKHS convolutional signal models.

One of the most promising aspects of the convolutional RKHS model on groups, is that it offers a natural way of performing discrete convolutions with different degrees of approximation. In particular, using RKHS convolutional models, there is no need to perform the so-called lifting of the signal on the group or to discretize the domain of the signals and the filters. Instead, RKHS convolutional models on groups are naturally endowed with representations whose discrete representation is embedded in the localization of the kernels when expanding the filters and the signals as in~\eqref{eqn_representation_property}. This provides a promising tool for the design of future invariant and equivariant machine learning architectures.

An interesting future research direction that opens as a byproduct of the use of RKHS algebras in convolutional neural networks is that of studying the rate of convergence of the gradient descent methods involved in the minimization of the cost function used to find the optimal weights in the network. In particular, it will be interesting to understand how the product operation in the algebra and the pointwise nonlinearity in the RKHS network affect the rate of convergence of gradient descent methods.

A real-life experiment in which we aim to predict a wireless coverage signal on a soccer field shows how  RKHS convolutional filters and AlgNNs can accommodate non-uniformly and varying sampling of the input signals, which are sampled along the stochastic trajectories of a UAV. 

One aspect that is worth exploring in future research efforts is the role of different configurations of filters, layers, and extra operators when building the RKHS based AlgNN. This will contribute to a deeper understanding of the computational limits and strengths of more general architectures based on RKHS representations.

%
%
%
%
%
%
%
%


\appendices



\section{Proof of Proposition~\ref{prop_starconv_vs_rkhsconv}}
\label{sec_proof_thm_Eucld_conv}

\begin{proof}
The convolution ``$\ast$", between a signal $f(x)=\sum_{v\in\ccalV}\alpha_{v}k_{v}(x)$ and a filter $g(x)=\sum_{u\in\ccalU}\beta_{u}k_{u}(x)$ results in $h\in\ccalH$ given by
\begin{equation}
h(x)=
\left(f \ast g \right)(x)
     = 
      \sum_{\substack{ v\in\ccalV \\ u\in\ccalU}}
            \alpha_{v}
            \beta_{u}
            \frac{B}{\pi}
            \sinc\left(
                       \frac{B}{\pi}
                           \left(
                              x - \left(v+u\right)
                           \right)
                  \right)
                  ,
\end{equation}
where $\ccalV,\ccalU\subset\mbR$. Now, taking into account~\eqref{eq_star_conv_sinc}, it follows that
\begin{equation}
h(x)
     = 
      \sum_{\substack{ v\in\ccalV \\ u\in\ccalU}} 
            \alpha_{v}
            \beta_{u}
                    \left(
                          \frac{B}{\pi}
                     \right)^{2}     
\sinc\left(
           \frac{B}{\pi}
                \left( 
                    x-v
                \right)    
     \right)
\star
\sinc\left(
           \frac{B}{\pi}
                \left( 
                    x-u
                \right)    
     \right)
          .
\end{equation}
Since ``$\star$" is a linear operator, we can re-group and rearrange terms to obtain
\begin{equation}
h(x)
      = 
        \left(
               \sum_{v\in\ccalV}
                          \alpha_{v}
                                     \sinc\left(
                                                   \frac{B}{\pi}
                                                    \left( 
                                                           x-v
                                                    \right)    
                                               \right)
        \right)   
        \star
        \left(
               \sum_{u\in\ccalU}
                          \beta_{u}
                                     \sinc\left(
                                                   \frac{B}{\pi}
                                                    \left( 
                                                           x-u
                                                    \right)    
                                               \right)
        \right)
        ,
\end{equation}
which leads to $h(x)=\left( f\star g\right)(x)$. Up to this point we have proved that ``$\ast$" and ``$\star$" are equivalent when considering signals in $\ccalS_{\ccalH}$. In what follows, we extend this proof to signals in $\ccalH\setminus\ccalS_{\ccalH}$.

As shown in Proposition 2.1 in~\cite[p.~17]{paulsen2016introduction} the span of the set of linear combinations of the form $\sum_{v\in\ccalV}\alpha_{v}k_{v}(x)$ is dense in $\ccalH$. This guarantees that for any $\epsilon>0$ and $f\in\ccalH$ there exists $f_{\epsilon}=\sum_{v\in\ccalV_{\epsilon}}\alpha_{v}(\epsilon)k_{v}(x)\in\ccalH$ and $E_{f,f_{\epsilon}}\in\ccalH$ such that
$ f = f_{\epsilon} + E_{f,f_{\epsilon}}$ with $\Vert E_{f,f_{\epsilon}}\Vert_{L_{2}}<\epsilon$. With this notion at hand, let us consider $\epsilon>0$ and $f,g\in\ccalH$. Then, we can write
\begin{equation}
f_{\epsilon}\star g_{\epsilon}
   =
     \left( 
           f
           -
           E_{f,f_{\epsilon}}
     \right)
     \star
     \left(
           g
           -
           E_{g,g_{\epsilon}}
     \right)
     .
\end{equation}
Distributing the product and organizing terms we obtain
\begin{equation}
f\star g
-
f_{\epsilon}\star g_{\epsilon}
   =
      f\star E_{g,g_{\epsilon}}
      +
      E_{f,f_{\epsilon}}\star g
      -
      E_{f,f_{\epsilon}}\star E_{g,g_{\epsilon}}
     .
\end{equation}
Now, taking into account that $f_{\epsilon}\star g_{\epsilon} = f_{\epsilon}\ast g_{\epsilon}$ it follows that
\begin{equation}
f\star g
-
f_{\epsilon}\ast g_{\epsilon}
   =
      f\star E_{g,g_{\epsilon}}
      +
      E_{f,f_{\epsilon}}\star g
      -
      E_{f,f_{\epsilon}}\star E_{g,g_{\epsilon}}
     .
\end{equation}
Now, taking the $L_{2}$-norm on both sides and using the triangular inequality we reach
\begin{multline}\label{eq_proof_con_star_ast_1}
\left\Vert
       f\star g
       -
       f_{\epsilon}\ast g_{\epsilon}
\right\Vert_{L_{2}}
   \leq
   \\
         \left\Vert  
               f\star E_{g,g_{\epsilon}}
         \right\Vert_{L_{2}}
         +
         \left\Vert 
                E_{f,f_{\epsilon}}\star g
         \right\Vert_{L_{2}}
         +
         \left\Vert   
                E_{f,f_{\epsilon}}\star E_{g,g_{\epsilon}}
         \right\Vert_{L_{2}}
     .
\end{multline}
From Young's inequality for the $\star$ convolution -- see~\cite{bogachev2007measure} -- we know that
\begin{equation}
\left\Vert 
       f
       \star 
       g
\right\Vert_{L_{r}}
               \leq
                \Vert 
                        f
                \Vert_{L_{p}}
                \Vert 
                        g
                \Vert_{L_{q}}
                ,
\end{equation}
where $1/p+1/q=1+1/r$. Then, taking into account this property for each term on the right hand side of~\eqref{eq_proof_con_star_ast_1} and the inequality $\Vert E_{f,f_{\epsilon}}\Vert_{L_{2}}<\epsilon$ we have that
\begin{equation}
\left\Vert
       f\star g
       -
       f_{\epsilon}\ast g_{\epsilon}
\right\Vert_{L_{2}}
   \leq
   \\
         \left\Vert  
               f
         \right\Vert_{L_{1}}
         \epsilon
         +
         \left\Vert 
                 g
         \right\Vert_{L_{1}}
         \epsilon
         +
         \left\Vert   
                E_{f,f_{\epsilon}}
         \right\Vert_{L_{1}}
         \epsilon
     .
\end{equation}
Since $f,g,E_{f,f_{\epsilon}}\in L_{1}(\mbR)$ and $\epsilon>0$ is arbitrary the proof is completed.

\end{proof}


\section{Proof of Proposition~\ref{prop_WsqW}}
\label{proof_prop_WsqW}

\begin{proof}

Any graphon $W: [0,1]^{2} \to [0,1]$ is a bounded symmetric measurable function, then from the application of Proposition 11.2. in~\cite[p.~147]{paulsen2016introduction} we have that $K=W\square W$ is an RKHS kernel function. Therefore, by Theorem 2.14~\cite[p.~25]{paulsen2016introduction} the kernel $K$ induces an RKHS $\ccalH\subset L_{2}([0,1])$.

\end{proof}


\section{Proof of Theorem~\ref{thm_general_A_rkhs}}
\label{proof_thm_general_A_rkhs}

\begin{proof}
We proceed to show that $\ccalA_{\ccalH}$ is an unital algebra. Since the sum of two countable sums is countable, $\ccalA_{\ccalH}$ is closed under the ordinary sum. Additionally, for any $\gamma\in\mbF$ and $\sum_{v\in\ccalV}\alpha_{v}k_{v}(t)\in\ccalA_{\ccalH}$
we have
\begin{equation}
 \gamma 
 \sum_{v\in\ccalV}\alpha_{v}k_{v}(t)
 =
 \sum_{v\in\ccalV}\gamma\alpha_{v}k_{v}(t)\in\ccalA_{\ccalH}
 .
\end{equation}
Then, it follows that $\ccalA_{\ccalH}$ is a vector space. Now, we turn our attention to the product operation. To formally show that the operation ``$\ast$" in~\eqref{eq_A_rkhs_gen_prod} is an algebra product, it must be true that~\cite{repthybigbook}:
\begin{equation}
\begin{tikzcd} 
\ccalA_{\ccalH}
\otimes
\ccalA_{\ccalH}
\otimes
\ccalA_{\ccalH}
\arrow[r,"\ast\otimes\text{Id}"]\arrow[d,"\text{Id}\otimes\ast"] &
\ccalA_{\ccalH}
\otimes
\ccalA_{\ccalH}
\arrow[d,"\ast"]
\\
\ccalA_{\ccalH}
\otimes
\ccalA_{\ccalH}
\arrow[r,"\ast"] & |[black, rotate=0]| 
\ccalA_{\ccalH}
\end{tikzcd}
,
\end{equation}
which implies
\begin{equation}
f\ast g \in\ccalA_{\ccalH},
\quad
\left(
f\ast g
\right)\ast
      h
=
f\ast
\left(
     g\ast h
\right)
,
\end{equation}
for all $f,g,h\in\ccalA_{\ccalH}$. The fact that $f\ast g \in\ccalA_{\ccalH}$ follows trivially from the definition of ``$\ast$". Now, let us consider
\begin{equation}
f
=
\left( 
     \sum_{v\in\ccalV}\alpha_{v}k_{v}(t)
\right)
,
~
g
=
\left( 
     \sum_{u\in\ccalU}\beta_{u}k_{u}(t)
\right)
,
h
=
\left( 
     \sum_{\ell\in\ccalL}\gamma_{\ell}k_{\ell}(t)
\right)
.
\end{equation}
Then, we start computing
\begin{equation}
 \left(
        f\ast g
\right)
\ast
     h
      =
         \sum_{v,u,\ell}
                       \alpha_{v}\beta_{u}\gamma_{\ell}
                       k_{\left(v\circ u\right)\circ\ell}
  .
\end{equation}
Since
$
\left(
       v \circ u
\right)
         \circ
               \ell
             =
                v
                \circ
                       \left( 
                                   u\circ\ell
                           \right)
                    ,
$
it follows that
\begin{equation}
 \left(
      f\ast g
\right)
      \ast h
      =
         \sum_{v,u,\ell}
                      \alpha_{v}\beta_{u}\gamma_{\ell}
                      k_{v\circ\left( u\circ\ell\right)}
      =
f\ast
   \left(
         g
            \ast
         h
   \right)   
  .
\end{equation}

Now, we proceed to show that there exists a unit element, $u_{\ccalA}\in\ccalA_{\ccalH}$, such that 
\begin{equation}
f
    \ast
u_{\ccalA_{\ccalH}}   
=
u_{\ccalA_{\ccalH}}  
    \ast 
f
=
f
.
\end{equation}
We claim that $u_{\ccalA_{\ccalH}}=k_{\delta}$, where $\delta$ is given by~\eqref{eq_A_rkhs_gen_prod_unit_element}. To see this notice that
\begin{equation}
 k_{\delta}
       \ast
 k_{v}
          =
             k_{\delta \circ v}
          =
             k_{v\circ\delta}
          =
             k_{v}
 \quad
 \forall~v\in\ccalX
 .
\end{equation}
\end{proof}


\section{Proof of Theorem~\ref{thm_asm_from_rkhs}}
\label{sec_proof_thm_asm_from_rkhs}

\begin{proof}
To probe Theorem~\ref{thm_asm_from_rkhs}, we proceed to show that $\rho$ as stated in~\eqref{eq_thm_asm_from_rkhs_1} is a homomorphism from $\ccalA_{\ccalH}$ to $\text{End}(\ccalH)$. To this end we show that $\rho$ preserves the product in the algebra, turning it into the composition of linear operators in $\text{End}(\ccalH)$. If $h_{1}, h_{2}\in \ccalA_{\ccalH}$, then it follows that
\begin{equation}
\rho\left( 
             h_{1}
      \right)
              \left(
                             \rho\left(
                                           h_{2}
                                    \right)
                                           f
              \right)
              =
                h_{1} 
                        \ast
                               \left(
                                    h_{2} \ast f
                                \right)  
                                .  
\end{equation}
Since ``$\ast$" is associative -- by Theorem~\ref{thm_general_A_rkhs} --, we can write
\begin{equation}
h_{1} 
        \ast
             \left(
                    h_{2} \ast f
             \right)  
                      =
                         \left( h_{1} 
                                        \ast
                                 h_{2} 
                         \right)        
                                 \ast f 
             .  
\end{equation}
Then, taking into account the definition of $\rho$ and the fact that ``$\ast$" is also the product in the algebra $\ccalA_{\ccalH}$ we have
\begin{equation}
\left( h_{1} 
                  \ast
        h_{2} 
\right)        
         \ast f   
                =
                  \rho\left( 
                               h_{1}
                                     \ast
                               h_{2}      
                         \right)   
                                f 
                                .
\end{equation}
Therefore, we finally obtain
\begin{equation}
\rho\left( 
             h_{1}
      \right)
              \left(
                             \rho\left(
                                           h_{2}
                                    \right)
                                           f
              \right)
                         =
                           \rho\left( 
                                  h_{1}
                                          \ast
                                  h_{2}      
                           \right)   
                                  f 
                                 .
\end{equation}
\end{proof}


\section{Proof of Corollary~\ref{thm_density_AH_in_H}}
\label{sec_proof_thm_density_AH_in_H}

\begin{proof}
We start taking into account that $\left\Vert \rho(h) g \right\Vert_{\ccalH} \leq C_{h}\Vert g\Vert_{\ccalH}$ for all $g\in\ccalH$ with $C_{h}>0$. Then, the action of $\rho(h)$ on $ f- \sum_{v\in\ccalV_{\epsilon}}\alpha_{v}(\epsilon)k_{v}(x) $ satisfies that
\begin{equation}
\left\Vert 
       h\ast f- h\ast\sum_{v\in\ccalV_{\epsilon}}\alpha_{v}(\epsilon)k_{v}(x) 
\right\Vert_{\ccalH}
        <
          C_{h}
          \left\Vert 
                        f
                        - 
                        \sum_{v\in\ccalV_{\epsilon}}\alpha_{v}(\epsilon)k_{v}(x) 
          \right\Vert_{\ccalH}
          .
\end{equation}
Then, tanking into account~\eqref{eq_thm_density_AH_in_H_1} we complete the proof. 
\end{proof}

\section{Proof of Corollary~\ref{cor_general_A_rkhs_group}}
\label{proof_cor_general_A_rkhs_group}

\begin{proof}

Since any group product $\circ$ is associative by definition~\cite{hall_liealg,hall_quantum,terrasFG}, we have that
\begin{equation}
\left( v \circ u\right)
     \circ 
     \ell
=
v\circ \left(
          u
          \circ
          \ell
       \right)
       ,
\end{equation}
and therefore~\eqref{eq_A_rkhs_gen_prod_associative_law} is satisfied. Additionally, when $\delta$ is the identity element of the group we have
\begin{equation}
\delta\circ u 
          = 
             u\circ \delta 
                       = u\quad \forall~u\in\ccalX  
 ,
\end{equation}
which satisfies~\eqref{eq_A_rkhs_gen_prod_unit_element}. Then, in virtue of Theorem~\ref{thm_general_A_rkhs} and Theorem~\ref{thm_asm_from_rkhs} we have the convolutional ASM $\left( \ccalA_{\ccalH}, \ccalS, \rho \right)$ for signals in an RKHS on the group. Finally, taking into account Corollary~\ref{thm_density_AH_in_H} we complete the proof. 
\end{proof}


\section{Proof of Theorem~\ref{thm_continuity_nonlinearity}: Continuity of Point-wise Nonlinearity}
\label{sub_sec_cont_pointwisenonl}

\begin{proof}

Let us consider $f(t)=\alpha k_{v_1}(t)+\beta k_{v_2}(t)$ and $g(t)=(\alpha+\beta)k_{v_1}(t)$. We can see that
\begin{equation}
 \lim_{v_{2}\to v_{1}}
     f(t)
     =
     g(t)
     .
\end{equation}
Now, we evaluate the action of $\eta$ on $f$ and $g$, respectively, to obtain 
\begin{equation}
\eta\left(
        f(t)
    \right)
    =
 \frac{
       \sigma\left( 
                         \alpha k_{v_1}(v_1)
                         +
                         \beta k_{v_{2}}(v_1)
                      \right)
      }
      {
      k_{v_1}(v_1)
      +
      k_{v_1}(v_2)
      }
+     
 \frac{
         \sigma\left( 
                         \alpha k_{v_1}(v_2)
                         +
                         \beta k_{v_{2}}(v_2)
                      \right)
      }
      {
      k_{v_2}(v_1)
      +
      k_{v_2}(v_2)
      }
      ,   
\end{equation}
%
%
%
%
%
%
\begin{equation}
\eta
    \left( 
        g(t)
    \right)
=
\frac{
       \sigma\left( 
                         (\alpha 
                           +
                           \beta)
                           k_{v_1}(v_1)
                 \right)
      }
      {
       k_{v_1}(v_1)
      }
      .
\end{equation}
Then, taking the limit $v_{2}\to v_{1}$ for $\eta(f(t))$ we have
\begin{equation}
    \lim_{v_{2}\to v_{1}}
     \eta\left(  
             f(t)
         \right)
         =
            \frac{
                    \sigma\left( 
                         (\alpha 
                         +
                         \beta) k_{v_{1}}(v_1)
                      \right)
                    }
                    {
                     k_{v_{1}}(v_{1})
                    }
          =
            \eta(g(t))
            .         
\end{equation}
\end{proof}


\bibliography{bibliography}
\bibliographystyle{unsrt}


\clearpage
\newpage

\newpage


\section{Frechet}
\label{sec_appendix_frechet}

Aiming to optimize the filters of the proposed AlgNNs, we consider the basic architecture in Fig.~\ref{fig:simulation_network}, where we have an RKHS network with  learnable  filters in $\ccalA_{\ccalH}$. We have two layers. In the first layer, we have a total of $N_1$ filters, $\left\lbrace \boldsymbol{w}_{j}^{(1,0)} \right\rbrace_{j=1}^{N_1}$, each followed by a point-wise non-linearity, $\eta(\cdot)$. Then, the features obtained are fed into a second layer, where we have $N_2$ filters, $\left\lbrace \boldsymbol{w}_{j}^{(2,i)} \right\rbrace_{i=1}^{N_2}$, per each feature -- each $j$ -- coming from layer one. We add up the output of those $N_2$ filters -- per each feature coming from the first layer, i.e. each $j$ -- and then apply a pointwise nonlinearity, $\eta(\cdot)$. Then, we add all the resultant signals again to obtain the final output of the network, which belongs to $\ccalH$. For an input, $f$, the network produces an output, $\boldsymbol{y}$, given by
\begin{equation}\label{eq_input_output_rel_network}
f_{out}
=
\sum_{i=1}^{N_2}
  \eta\left(
         \sum_{j=1}^{N_1}
              \boldsymbol{w}_{j}^{(2,i)}
                   \ast
               \eta\left(
                       \boldsymbol{w}_{j}^{(1,0)}
                            \ast
                       f
                   \right)
   \right)
   .
\end{equation}

Then, our goal when training the network is to find the filters in~\eqref{eq_input_output_rel_network} that minimize the quadratic error, measured when comparing the output of the network to the given input and reference signals. To show this explicitly, let us rewrite~\eqref{eq_input_output_rel_network} as
\begin{equation}\label{eq_input_output_rel_network_2}
f_{out}
     =
      F\left(
           \boldsymbol{w}_T
       \right)
       ,
\end{equation}
with
\begin{equation}
\boldsymbol{w}_{T}
    =
\left(
    \left(
        \boldsymbol{w}_{j}^{1,0}
    \right)_{j=1}^{j=N_1}
    ,
    \left(
        \boldsymbol{w}_{j}^{2,i}
    \right)_{j=1,i=1}^{j=N_1,i=N_2}
\right)
.
\end{equation}

Then, given an input $f$ to the RKHS network and a reference signal $r\in \ccalH$ we aim to minimize
\begin{equation}\label{eq_ell_Fh}
\ell(\boldsymbol{w}_{T})
        =
         \frac{1}{2}
            \left\Vert
                  r
                  -
                    F\left(
                         \boldsymbol{w}_{T}
                      \right)
            \right\Vert_{\ccalH}^2
            .
\end{equation}
To write explicit expressions for gradient descent we require the calculation of the Fr\'echet derivative of
$
F\left(
      \boldsymbol{w}_T
  \right)
  .
$
We can achieve this by taking into account that
\begin{equation}
 F
 \left(
      \boldsymbol{w}_T
  \right)
     =
F^{(1)}
   \left(
       \boldsymbol{w}_{j}^{(1,0)}
   \right)
   =
F^{(2)}
   \left(
       \boldsymbol{w}_{j}^{(2,i)}
   \right)
   .
\end{equation}
This is, the expressions of $F(1)$ and $F^{(2)}$ coincide with that of $F$, but the terms $\boldsymbol{w}_{j}^{(2,i)}$ and $\boldsymbol{w}_{j}^{(1,0)}$  are considered as constants, respectively. Then, by means of the Fr\'echet derivative properties -- see~\cite{berger1977nonlinearity} pages 69-71 -- we have
\begin{multline}
\bbD_{F}\left(
           \boldsymbol{w}_{T}
        \right)
     \left\lbrace
          \boldsymbol{d}_{T}
     \right\rbrace
     =
     \sum_{j=1}^{N_1}\bbD_{F^{(1)}}
             \left(
                  \boldsymbol{w}_{j}^{(1,0)}
             \right)
                  \left\lbrace
                       \boldsymbol{d}_{j}^{(1,0)}
                  \right\rbrace
      +
      \\
       \sum_{j=1}^{N_1}
       \sum_{i=1}^{N_2}\bbD_{F^{(2)}}
             \left(
                  \boldsymbol{w}_{j}^{(2,i)}
             \right)
                  \left\lbrace
                       \boldsymbol{d}_{j}^{(2,i)}
                  \right\rbrace
                  ,
\end{multline}
where

\begin{equation}
\boldsymbol{d}_{T}
    =
\left(
    \left(
        \boldsymbol{d}_{j}^{(1,0)}
    \right)_{j=1}^{j=N_1}
    ,
    \left(
        \boldsymbol{d}_{j}^{(2,i)}
    \right)_{j=1,i=1}^{j=N_1,i=N_2}
\right)
.
\end{equation}

In the following Theorem, we derive explicit expressions for the Fr\'echet derivatives of $F^{(1)}$ and $F^{(2)}$.


\begin{theorem}\label{thm_F1_F2_derivatives}
Let $\ccalH$ be an RKHS with reproducing kernel $K(u,v)$ and let $F^{(1)}:\ccalH \to \ccalH$ and $F^{(2)}:\ccalH \to \ccalH$ be given by
\begin{multline}
 F^{(1)}\left(
      \boldsymbol{w}_{j}^{(1,0)}
  \right)
  =
\sum_{i=1}^{N_2}
  \eta\left(
         \sum_{j=1}^{N_1}
              \boldsymbol{w}_{j}^{(2,i)}
                   \ast
               \eta\left(
                       \boldsymbol{w}_{j}^{(1,0)}
                            \ast
                       f
                   \right)
   \right)
   ,
\end{multline}
and
\begin{multline}
 F^{(2)}\left(
            \boldsymbol{w}_{j}^{(2,i)}
        \right)
  =
\sum_{i=1}^{N_2}
  \eta\left(
         \sum_{j=1}^{N_1}
              \boldsymbol{w}_{j}^{(2,i)}
                   \ast
               \eta\left(
                       \boldsymbol{w}_{j}^{(1,j)}
                            \ast
                       f
                   \right)
   \right)
   .
\end{multline}

Then, it follows that
\begin{multline}
\bbD_{F^{(1)}}\left(
         \boldsymbol{w}_{j}^{(1,0)}
      \right)
     \left\lbrace
        \boldsymbol{d}_{j}^{(1,0)}
     \right\rbrace
     =
     \\
      \sum_{i=1}^{N_2}
            \bbD_{\eta}
                 \left(
                     \boldsymbol{w}_{j}^{(2,i)}
                        \ast
                      \boldsymbol{w}_{j}^{(1,0)}
                      +
                      \sum_{k\neq j}
                           \boldsymbol{w}_{k}^{(2,i)}
                           \ast
                           \eta\left(
                                  \boldsymbol{w}_{k}^{(1,0)}
                                    \ast
                                  f
                               \right)
                 \right)
                 \\
                    \left\lbrace
                        \boldsymbol{w}_{j}^{(2,i)}
                          \ast
                                 \bbD_{\eta}\left(
                       \boldsymbol{w}_{j}^{(1,0)}
                            \ast
                        f
                    \right)
                          \left\lbrace
                               \boldsymbol{d}_{j}^{(1,0)}
                               \ast
                               f
                          \right\rbrace
                    \right\rbrace
                    ,
\end{multline}
\begin{multline}
\bbD_{F^{(2)}}
          \left(
               \boldsymbol{w}_{j}^{(2,i)}
          \right)
    \left\lbrace
        \boldsymbol{d}_{j}^{(2,i)}
    \right\rbrace
    =
    \\
    \bbD_{\eta}\left(
                    \boldsymbol{w}_{j}^{(2,i)}
                       \ast
                    \eta\left(
                           \boldsymbol{w}_{j}^{(1,0)}
                              \ast
                           f
                        \right)
                    +
                    \sum_{k\neq j}
                        \boldsymbol{w}_{k}^{(2,i)}
                             \ast
                        \eta\left(
                                \boldsymbol{w}_{j}^{(1,0)}
                                   \ast
                                f
                            \right)
                \right)
                \\
      \left\lbrace
          \boldsymbol{d}_{j}^{(2,i)}
              \ast
          \eta\left(
                   \boldsymbol{w}_{j}^{(1,0)}
                      \ast
                   f
              \right)
      \right\rbrace
      .
\end{multline}
\end{theorem}

\begin{proof}
    See Appendix~\ref{proof_thm_F1_F2_derivatives}.
\end{proof}


In what follows we derive concrete expressions for the derivatives of the pointwise non-linearities, $\bbD_{\eta}(\boldsymbol{w})\{ \boldsymbol{d} \}$, taking into account the considerations discussed in Section~\ref{sub_sec_nonlinearity}.


\begin{theorem}\label{thm_frechet_nonlinearity}
Let $\ccalH$ be an RKHS with reproducing kernel $K(u,v)$ and let $\eta: \ccalH\to \ccalH$ be a point-wise non-linearity as specified in~\eqref{eq_nonlinearity_a}. Then, the Fr\'echet derivative of $\eta$ evaluated at
\begin{equation}
\displaystyle\boldsymbol{w}=
             \sum_{u\in\ccalU_1}\bbh(u)k_{u}
\end{equation}
and acting on
\begin{equation}
 \boldsymbol{d}=
          \sum_{u\in\ccalU_2}\bbd(u)k_{u}
\end{equation}
is given by
\begin{equation}\label{eq_thm_frechet_nonlinearity_1}
 \bbD_{\eta}(\boldsymbol{w})
          \left\lbrace
                 \boldsymbol{d}
          \right\rbrace
          =
          \sum_{u\in\ccalU}
                     \frac{
    \sigma^{'}\left(
                  \boldsymbol{w}(u)
              \right)
              \boldsymbol{d}(u)
    }
    {
     \sum_{r,u\in\ccalU}k_{u}(r)
    }
                 k_{u}
          ,
\end{equation}
where $\ccalU = \ccalU_1 \bigcup \ccalU_2$ and $\sigma^{'}$ is the ordinary derivative of $\sigma$.
\end{theorem}

\begin{proof}
    See Appendix~\ref{proof_thm_frechet_nonlinearity}
\end{proof}


Notice that the expression in~\eqref{eq_thm_frechet_nonlinearity_1} does not require $\boldsymbol{w}$ and $\boldsymbol{d}$ being represented with $\ccalU_1 = \ccalU_2$.

We now state the expression that defines the derivative of the cost function, $\ell(\boldsymbol{w})$, in terms of $F(\boldsymbol{w})$ in~\eqref{eq_ell_Fh}. This will be of essential importance to state the steepest descend formulation of the problem.


\begin{prop}\label{prop_derivative_loss_Fh}
Let $\ccalH$ be an RKHS with reproducing kernel $K$. Given a reference signal $r\in \mathcal H$, a filter $\boldsymbol{w}\in \ccalA_{\ccalH}$ -- see Theorem~\ref{thm_general_A_rkhs} --  and
$
\ell(
    \boldsymbol{w}
    )
    =\frac{1}{2}
         \left\Vert
               r
               -
               F\left(
                    \boldsymbol{w}
                \right)
         \right\Vert_{\ccalH}^2
,
$
it follows that
\begin{equation}
\bbD_{\ell}(\boldsymbol{w})
     \left\lbrace
           \boldsymbol{d}
     \right\rbrace
     =
     -
     \left\langle
           \bbD_{F}(\boldsymbol{w})\left\lbrace
                                        \boldsymbol{d}
                                   \right\rbrace
           ,
          r
           -
           F(\boldsymbol{w})
    \right\rangle_{\ccalH}
    ,
\end{equation}
where $F$ is Fr\'echet differentiable.
\end{prop}

\begin{proof}
    See Appendix~\ref{proof_prop_derivative_loss_Fh}
\end{proof}



\subsubsection{Steepest descent formulation}
By leveraging the results in Proposition~\ref{prop_derivative_loss_Fh} we propose a steepest descend algorithm to compute the value of the filters $\boldsymbol{w}$ that minimize the cost function
$
\ell(
    \boldsymbol{w}
    )
    =\frac{1}{2}
         \left\Vert
               r
               -
               F\left(
                    \boldsymbol{w}
                \right)
         \right\Vert_{\ccalH}^2
.
$
To show this we start recalling that given $\ell(\boldsymbol{w})$ and its Fr\'echet derivative we have
\begin{equation}
\ell\left(
        \boldsymbol{w}_{k}
        +
        \alpha
        \boldsymbol{d}
    \right)
=
\ell(\boldsymbol{w}_{k})
+
\bbD_{\ell}(\boldsymbol{w}_{k})
   \left\lbrace
        \alpha\boldsymbol{d}
   \right\rbrace
+
o\left(
     \boldsymbol{w}
 \right)
 ,
\end{equation}
which in the light of Proposition~\ref{prop_derivative_loss_Fh} leads to
\begin{multline}
\ell\left(
        \boldsymbol{w}_{k}
        +
        \alpha
        \boldsymbol{d}
    \right)
=
\\
\ell(\boldsymbol{w}_{k})
     -
     \alpha
     \left\langle
           \bbD_{F}(\boldsymbol{w}_{k})\left\lbrace
                                        \boldsymbol{d}
                                   \right\rbrace
           ,
           r
           -
           F(\boldsymbol{w}_{k})
    \right\rangle_{\ccalH}
+
o\left(
     \boldsymbol{w}
 \right)
 .
\end{multline}
Then, to find the direction, $\boldsymbol{d}$, of the fastest decrease we must select $\boldsymbol{d}$ as the solution to the following problem.
\begin{equation}\label{eq_opt_problm_aux_1}
   \sup_{\boldsymbol{d}}
       \left\langle
           \bbD_{F}(\boldsymbol{w}_{k})\left\lbrace
                                        \boldsymbol{d}
                                   \right\rbrace
           ,
           r
           -
           F(\boldsymbol{w}_{k})
    \right\rangle_{\ccalH}
    ~
    \text{s.t.}
    ~
    \Vert \boldsymbol{d}\Vert_{\ccalH}=1
    .
\end{equation}
The solution of~\eqref{eq_opt_problm_aux_1} is achieved when $\boldsymbol{d}$ is selected such that
\begin{equation}\label{eq_inner_prod_eqmax}
 \bbD_{F}(\boldsymbol{w}_{k})
     \left\lbrace
          \boldsymbol{d}
     \right\rbrace
     =
      r
      -
      F(\boldsymbol{w}_{k})
      .
\end{equation}
If we denote the solution of~\eqref{eq_inner_prod_eqmax} by $\tilde{\boldsymbol{d}}_{k}$, then we have that the values of $\boldsymbol{w}$ in the $k$-step of an iterative search is given by
\begin{equation}\label{eq_iter_hk}
\boldsymbol{w}_{k+1}
=
\boldsymbol{w}_{k}
+
\alpha_{k}
\boldsymbol{d}_{k}
,
\end{equation}
where
$
\boldsymbol{d}_{k} = \tilde{\boldsymbol{d}}_{k} / \Vert \tilde{\boldsymbol{d}}_{k} \Vert_{\ccalH}
,
$
and where the values of $\alpha_{k}$ can be selected according to the so called Wolfe conditions~\cite{nocedal2006numerical}.

In Algorithms~\ref{alg_get_d} and~\ref{alg_alphak_update_rule} we make explicit how to solve ~\eqref{eq_inner_prod_eqmax} using a conjugate gradient descent and how to select the values of $\alpha_{k}$ in~\eqref{eq_iter_hk}.


\begin{algorithm}
\caption{
Finding $\boldsymbol{d}$ in
$
\bbD_{F}(\boldsymbol{w})
     \left\lbrace
          \boldsymbol{d}
     \right\rbrace
=
    r
    -
    F(\boldsymbol{w})
    .
$
}\label{alg_get_d}
\begin{algorithmic}
\Require $\boldsymbol{w}$, $F(\boldsymbol{w})$, $r$, $\bbD_{F}(\boldsymbol{w})\{ \cdot \}$
\Ensure $\boldsymbol{d}$
\State $\boldsymbol{d}_{0} = 0$
\State $\boldsymbol{s}_{0} = r
    -
    F(\boldsymbol{w})
    -
    \bbD_{F}(\boldsymbol{w})
     \left\lbrace
          \boldsymbol{d}_{0}
     \right\rbrace
    $
\State $\boldsymbol{p}_{0} = \boldsymbol{s}_{0}$
\State $k=0$
\State $\texttt{aux}=1$
\While{$\texttt{aux}=1$}
\State
$\displaystyle
\gamma_{k}
   =
   \frac{
        \left\langle
               \boldsymbol{s}_{k}
               ,
               \boldsymbol{s}_{k}
        \right\rangle_{\ccalH}
        }
        {
        \left\langle
               \boldsymbol{p}_{k}
               ,
               \bbD_{F}(\boldsymbol{w})\{ \boldsymbol{p}_{k} \}
        \right\rangle_{\ccalH}
        }
$
\State $\boldsymbol{d}_{k+1} = \boldsymbol{d}_{k} + \gamma_{k}\boldsymbol{p}_{k
}$
\State $\boldsymbol{s}_{k+1} = \boldsymbol{s}_{k} - \gamma_{k}\bbD_{F}(\boldsymbol{w})\{ \boldsymbol{p}_{k} \}$
\If{$\Vert \boldsymbol{s}_{k+1}\Vert_{\ccalH}<\epsilon$}
    \State $\boldsymbol{d}=\boldsymbol{d}_{k+1}$
    \State $\texttt{aux}=0$
\Else
    \State
$\displaystyle
\beta_{k}
   =
   \frac{
        \left\langle
               \boldsymbol{s}_{k+1}
               ,
               \boldsymbol{s}_{k+1}
        \right\rangle_{\ccalH}
        }
        {
        \left\langle
               \boldsymbol{s}_{k}
               ,
               \boldsymbol{s}_{k}
        \right\rangle_{\ccalH}
        }
$
\State $\boldsymbol{p}_{k+1}=\boldsymbol{s}_{k+1}+\beta_{k}\boldsymbol{p}_{k}$
\State $k=k+1$
\EndIf
\EndWhile
\end{algorithmic}
\end{algorithm}



\begin{algorithm}
\caption{Update rule for $\alpha_k$ in~\eqref{eq_iter_hk}}\label{alg_alphak_update_rule}
\begin{algorithmic}
\Require $\overline{\alpha}>0$, $\rho\in (0,1)$, $c\in (0,1)$
\Ensure $\alpha_k$
\State $\alpha = \overline{\alpha}$
\State $\texttt{aux}=1$
\While{$\texttt{aux}=1$}
\If{
\begin{equation*}
\boldsymbol{\ell}\left(
                    \boldsymbol{w}_{k}
                    +
                    \alpha
                    \boldsymbol{d}_{k}
                 \right)
                 \leq
\boldsymbol{\ell}\left(
                     \boldsymbol{w}_{k}
                 \right)
                 -
                 c
                 \alpha
                 \left\langle
           \bbD_{F}(\boldsymbol{w}_{k})\left\lbrace
                                        \boldsymbol{d}_{k}
                                   \right\rbrace
           ,
           r
           -
           F(\boldsymbol{w}_{k})
    \right\rangle_{\ccalH}
\end{equation*}
}
\State $\texttt{aux}=0$
\Else
   \State $\alpha \gets \rho \alpha$
\EndIf
\EndWhile
\State $\alpha_{k}= \alpha$
\end{algorithmic}
\end{algorithm}


\section{Fr\'echet Derivatives of Cost Function}\label{sec_frechet_derivatives}


\begin{theorem}\label{thm_Frcht_comp}
Let $\ccalH_1$, $\ccalH_2$ and $\ccalH_3$ be normed vector spaces and let $\ccalU\subset\ccalH_1$ be an open subset of $\ccalH_1$. Let us consider the functions $F_{1}: \ccalH_1 \to \ccalH_2$, $F_2: \ccalH_2 \to \ccalH_3$ and the composition between $F_1$ and $F_2$ represented by $F_{2}\circ F_{1}: \ccalH_1 \to \ccalH_3$. Let $\bbD_{F_1}(u)\{ \cdot \}$ and $\bbD_{F_2}(F_{1}(\boldsymbol{u}))\{ \cdot \}$ be the Fr\'echet derivatives of $F$ and $F_2$ at $u$ and $F_{1}(\boldsymbol{u})$, respectively. Then we have
\begin{equation}
\bbD_{\left(F_{2}\circ F_{1}\right)(\boldsymbol{u})}  \{ \boldsymbol{w} \}
=
\bbD_{F_{2}}(F_{1}(\boldsymbol{u})) \left\lbrace
                        \bbD_{F_{1}}(\boldsymbol{u})
                              \{ \boldsymbol{w} \}
                    \right\rbrace
                    .
\end{equation}
\end{theorem}

\begin{proof}
    See Appendix~\ref{proof_thm_Frcht_comp}.
\end{proof}

\begin{prop}\label{prop_derivative_nonlin_and_filter}
Let $\ccalH$ be an RKHS with reproducing kernel $K$. Given a filter $\boldsymbol{w}\in \ccalA_{\ccalH}$ -- see Theorem~\ref{thm_general_A_rkhs} --  and an input signal $f\in \mathcal H$, let $F(\boldsymbol{w}):\mathcal H\to \ccalH$ as
$
F (\boldsymbol{w})
      =
        \eta\left(
               \boldsymbol{w}\ast f
            \right)
$,
where $\eta$ is a point-wise non-linearity as specified in Section~\ref{sub_sec_nonlinearity}. Then,
\begin{equation}
   \bbD_{F}(\boldsymbol{w})
         \left\lbrace
               \boldsymbol{d}
         \right\rbrace
         =
         \bbD_{\eta}\left(
                       \boldsymbol{w}\ast f
                    \right)
                          \left\lbrace
                               \boldsymbol{d}
                               \ast
                               f
                          \right\rbrace
                          .
\end{equation}
\end{prop}

\begin{proof}
    See Appendix~\ref{proof_prop_derivative_nonlin_and_filter}.
\end{proof}

\subsection{Proof of Theorem~\ref{thm_F1_F2_derivatives}}
\label{proof_thm_F1_F2_derivatives}

\begin{proof}

First, we start considering the functions
\begin{equation}
G_{1} (\boldsymbol{w})
     =
     \eta\left(
              \boldsymbol{w}\ast\boldsymbol{\alpha}
         \right)
         ,
\end{equation}
and
\begin{equation}
G_{2} (\boldsymbol{w})
     =
     \sum_{i=1}^{N_2}
          \eta
             \left(
                  \boldsymbol{\beta}_i
                    \ast
                  \boldsymbol{w}
                  +
                  \boldsymbol{\gamma}_i
             \right)
             .
\end{equation}

Then, we take into account that $F^{(1)}(\boldsymbol{w}) = \left( G_{2}\circ G_1 \right) (\boldsymbol{w})$. If we apply the chain rule for the Fr\'echet derivative of
$\left( G_{2}\circ G_1 \right) (\boldsymbol{w})$ it follows that
\begin{equation}
\bbD_{\left(G_2 \circ G_1 \right)}(\boldsymbol{w})\{ \boldsymbol{d} \}
=
\bbD_{G_2}\left(
               G_1 (\boldsymbol{w})
           \right)
\left\lbrace
                \bbD_{G_1}(\boldsymbol{w})\{ \boldsymbol{d} \}
                    \right\rbrace
\end{equation}
where we know that
\begin{equation}
   \bbD_{G_1}(\boldsymbol{w})
         \left\lbrace
               \boldsymbol{d}
         \right\rbrace
         =
         \bbD_{\eta}\left(
                       \boldsymbol{w}\ast\boldsymbol{\alpha}
                    \right)
                          \left\lbrace
                               \boldsymbol{d}
                               \ast
                               \boldsymbol{\alpha}
                          \right\rbrace
                          ,
\end{equation}
and
\begin{equation}
 \bbD_{G_2}(\boldsymbol{w})
      \{ \boldsymbol{d} \}
      =
      \sum_{i=1}^{N_2}
            \bbD_{\eta}
                 \left(
                     \boldsymbol{\beta}_i
                        \ast
                      \boldsymbol{w}
                      +
                      \boldsymbol{\gamma}_i
                 \right)
                    \left\lbrace
                        \boldsymbol{\beta}_i
                          \ast
                        \boldsymbol{d}
                    \right\rbrace
                    .
\end{equation}
Therefore,
\begin{multline}
\bbD_{F^{(1)}}(\boldsymbol{w})\{ \boldsymbol{d} \}
     =
     \\
      \sum_{i=1}^{N_2}
            \bbD_{\eta}
                 \left(
                     \boldsymbol{\beta}_i
                        \ast
                      \boldsymbol{w}
                      +
                      \boldsymbol{\gamma}_i
                 \right)
                    \left\lbrace
                        \boldsymbol{\beta}_i
                          \ast
                                 \bbD_{\eta}\left(
                       \boldsymbol{w}\ast\boldsymbol{\alpha}
                    \right)
                          \left\lbrace
                               \boldsymbol{d}
                               \ast
                               \boldsymbol{\alpha}
                          \right\rbrace
                    \right\rbrace
                    .
\end{multline}

Now, we evaluate the expressions above for $\boldsymbol{w}=\boldsymbol{w}_{j}^{(1,0)}$, i.e.
$
F\left(
    \boldsymbol{w}_{j}^{(1,0)}
  \right)
     =
      \left(
       G_{2} \circ G_{1}
      \right)\left(
                \boldsymbol{w}_{j}^{(1,0)}
             \right)
             ,
$
which leads to
%
%
%

%
%
%
\begin{multline}
\bbD_{F}\left(
         \boldsymbol{w}_{j}^{(1,0)}
      \right)
     \left\lbrace
          \boldsymbol{d}_{j}^{(1,0)}
     \right\rbrace
     =
     \\
      \sum_{i=1}^{N_2}
            \bbD_{\eta}
                 \left(
                     \boldsymbol{w}_{j}^{(2,i)}
                        \ast
                      \boldsymbol{w}_{j}^{(1,0)}
                      +
                      \sum_{k\neq j}
                           \boldsymbol{w}_{k}^{(2,i)}
                           \ast
                           \eta\left(
                                  \boldsymbol{w}_{k}^{(1,0)}
                                    \ast
                                  f
                               \right)
                 \right)
                 \\
                    \left\lbrace
                        \boldsymbol{w}_{j}^{(2,i)}
                          \ast
                                 \bbD_{\eta}\left(
                       \boldsymbol{w}_{j}^{(1,0)}
                            \ast
                        f
                    \right)
                          \left\lbrace
                               \boldsymbol{d}_{j}^{(1,0)}
                               \ast
                               f
                          \right\rbrace
                    \right\rbrace
\end{multline}

Now, we turn our attention to the Fr\'echet derivative of $F^{(2)}$. We start taking into account that
$
F^{(2)}(\boldsymbol{w})
 =
  \left(
     G_{2} \circ G_{1}
  \right)(\boldsymbol{w})
$
where
\begin{equation}
G_{1} (\boldsymbol{w})
     =
     \boldsymbol{w}
        \ast
      \boldsymbol{\alpha}
      +
      \boldsymbol{\beta}_{i,j}
      ,
\end{equation}
and
\begin{equation}
G_{2} (\boldsymbol{w})
     =
         \eta\left(
               \boldsymbol{w}
              \right)
         +
         \boldsymbol{\gamma}_{i,j}
         .
\end{equation}

If we take into account the chain rule property of the Fr\'echet derivative, it follows that
\begin{equation}
\bbD_{\left( G_2 \circ G_1 \right)}(\boldsymbol{w})\{ \boldsymbol{d} \}
=
\bbD_{G_2}\left(
               G_1 (\boldsymbol{w})
           \right)
\left\lbrace
                \bbD_{G_1}(\boldsymbol{w})\{ \boldsymbol{d} \}
                    \right\rbrace
                    ,
\end{equation}
where
\begin{equation}
\bbD_{G_1}(\boldsymbol{w})
    \left\lbrace
        \boldsymbol{d}
    \right\rbrace
    =
    \boldsymbol{d}
       \ast
    \boldsymbol{\alpha}
    ,
\end{equation}
and
\begin{equation}
\bbD_{G_2}(\boldsymbol{w})
    \left\lbrace
        \boldsymbol{d}
    \right\rbrace
    =
    \bbD_{\eta}(\boldsymbol{w})
      \left\lbrace
          \boldsymbol{d}
      \right\rbrace
      .
\end{equation}
Therefore,
\begin{equation}
\bbD_{\left( G_2 \circ G_1 \right)}(\boldsymbol{w})
    \left\lbrace
        \boldsymbol{d}
    \right\rbrace
    =
    \bbD_{\eta}\left(
                    \boldsymbol{w}
                       \ast
                    \boldsymbol{\alpha}
                    +
                    \boldsymbol{\beta}_{i,j}
                \right)
      \left\lbrace
          \boldsymbol{d}
              \ast
          \boldsymbol{\alpha}
      \right\rbrace
      .
\end{equation}

Then, evaluating the expressions above in $\boldsymbol{w}=\boldsymbol{w}_{j}^{(2,i)}$, it follows that
\begin{multline}
\bbD_{F^{(2)}}
          \left(
               \boldsymbol{w}_{j}^{(2,i)}
          \right)
    \left\lbrace
        \boldsymbol{d}_{j}^{(2,i)}
    \right\rbrace
    =
    \\
    \bbD_{\eta}\left(
                    \boldsymbol{w}_{j}^{(2,i)}
                       \ast
                    \eta\left(
                           \boldsymbol{w}_{j}^{(1,0)}
                              \ast
                           f
                        \right)
                    +
                    \sum_{k\neq j}
                        \boldsymbol{w}_{k}^{(2,i)}
                             \ast
                        \eta\left(
                                \boldsymbol{w}_{j}^{(1,0)}
                                   \ast
                                f
                            \right)
                \right)
                \\
      \left\lbrace
          \boldsymbol{d}_{j}^{(2,i)}
              \ast
          \eta\left(
                   \boldsymbol{w}_{j}^{(1,0)}
                      \ast
                   f
              \right)
      \right\rbrace
\end{multline}

\end{proof}


\subsection{Proof of Theorem~\ref{thm_frechet_nonlinearity}}
\label{proof_thm_frechet_nonlinearity}

\begin{proof}

We start by leveraging the definition of the Fr\'echet derivative of $\eta$.
\begin{multline}
\eta\left(
         \boldsymbol{w}
         +
         \boldsymbol{d}
    \right)
    -
\eta\left(
         \boldsymbol{w}
    \right)
    =
    \\
 \eta\left(
         \sum_{u\in\ccalU_1}\bbh(u)k_{u}
         +
         \sum_{u\in\ccalU_2}\bbd(u)k_{u}
    \right)
    -
\eta\left(
         \sum_{u\in\ccalU_1}\bbh(u)k_{u}
    \right)
    .
\end{multline}
Now, we take into account that if $\ccalU = \ccalU_1 \bigcup \ccalU_2$ we can write
\begin{equation}
 \boldsymbol{w}=\sum_{u\in\ccalU}\widetilde{\bbh}(u)k_{u}
 ,
 \quad
 \boldsymbol{d}=\sum_{u\in\ccalU}\widetilde{\bbd}(u)k_{u}
\end{equation}
where
\begin{equation}
\widetilde{\bbh}(u)=
\left\lbrace
     \begin{array}{ccc}
         \bbh(u)  &  \text{if} & u\in\ccalU_{1} \\
         0 &     \text{otherwise} &
     \end{array}
\right.
,
\end{equation}
and
\begin{equation}
\widetilde{\bbd}(u)=
\left\lbrace
     \begin{array}{ccc}
         \bbd(u)  &  \text{if} & u\in\ccalU_{2} \\
         0 &     \text{otherwise} &
     \end{array}
\right.
.
\end{equation}

Applying the action of $\eta$ in terms of $\widehat{\eta}$ -- see~\eqref{eq_nonlinearity_a} -- we obtain
\begin{multline}
\eta\left(
         \boldsymbol{w}
         +
         \boldsymbol{d}
    \right)
    -
\eta\left(
         \boldsymbol{w}
    \right)
    =
    \\
    \sum_{u\in\ccalU}\widehat{\eta}_{u}
                \left(
                      \boldsymbol{w}
                      +
                      \boldsymbol{d}
                 \right)
                       k_{u}
    -
    \sum_{u\in\ccalU}
          \widehat{\eta}_{u}
          \left(
              \boldsymbol{w}
          \right)
          k_{u}
          .
\end{multline}
Factoring out the term $k_{u}$ in the sum we obtain
\begin{equation}
\eta\left(
         \boldsymbol{w}
         +
         \boldsymbol{d}
    \right)
    -
\eta\left(
         \boldsymbol{w}
    \right)
    =
         \sum_{u\in\ccalU}
         \left(
         \widehat{\eta}_{u}\left(
                      \boldsymbol{w}
                      +
                      \boldsymbol{d}
                 \right)
    -
          \widehat{\eta}_{u}
          \left(
              \boldsymbol{w}
          \right)
          \right)
          k_{u}
          ,
\end{equation}
which implies
\begin{equation}
 \bbD_{\eta}(\boldsymbol{w})
          \left\lbrace
                 \boldsymbol{d}
          \right\rbrace
          =
          \sum_{u\in\ccalU}
                      \bbD_{\widehat{\eta}_{u}}
                           \left(
                               \boldsymbol{w}
                           \right)
                           \left\lbrace
                                  \boldsymbol{d}
                           \right\rbrace
                 k_{u}
                 .
\end{equation}

Now, we calculate the explicit expression for
$ \bbD_{\widehat{\eta}_{u}}
                           \left(
                               \boldsymbol{w}
                           \right)
                           \left\lbrace
                                  \boldsymbol{d}
                           \right\rbrace
                           .
$
We proceed taking into account that
\begin{equation}
\widehat{\eta}_{u}\left(
         \boldsymbol{w}
         +
         \boldsymbol{d}
    \right)
    -
\widehat{\eta}_{u}\left(
         \boldsymbol{w}
    \right)
    =
    \frac{
       \sigma\left( \boldsymbol{w}(u)+ \boldsymbol{d}(u)\right)
      }
      {
      \sum_{r,u\in\ccalU}k_{u}(r)
      }
      -
      \frac{
       \sigma\left(\boldsymbol{w}(u)\right)
      }
      {
      \sum_{r,u\in\ccalU}k_{u}(r)
      }
          .
\end{equation}
Then, we have
\begin{equation}
\widehat{\eta}_{u}\left(
         \boldsymbol{w}
         +
         \boldsymbol{d}
    \right)
    -
\widehat{\eta}_{u}\left(
         \boldsymbol{w}
    \right)
    =
    \frac{
    \sigma^{'}\left(
                  \boldsymbol{w}(u)
              \right)
              \boldsymbol{d}(u)
    }
    {
     \sum_{r,u\in\ccalU}k_{u}(r)
    }
    =
                          \bbD_{\widehat{\eta}_{u}}
                           \left(
                               \boldsymbol{w}
                           \right)
                           \left\lbrace
                                  \boldsymbol{d}
                           \right\rbrace
                           .
\end{equation}
With this expression at hand, we finally have
\begin{equation}
 \bbD_{\eta}(\boldsymbol{w})
          \left\lbrace
                 \boldsymbol{d}
          \right\rbrace
          =
          \sum_{u\in\ccalU}
                     \frac{
    \sigma^{'}\left(
                  \boldsymbol{w}(u)
              \right)
              \boldsymbol{d}(u)
    }
    {
     \sum_{r,u\in\ccalU}k_{u}(r)
    }
                 k_{u}
          ,
\end{equation}
where $\sigma^{'}(\cdot)$ is the ordinary derivative of $\sigma(x)=\max\{ 0,x \}$.

\end{proof}

\subsection{Proof of Proposition~\ref{prop_derivative_loss_Fh}}
\label{proof_prop_derivative_loss_Fh}

\begin{proof}

First, we start taking into account that
\begin{equation}
\ell\left(
         \boldsymbol{w}
    \right)
    =
    \frac{1}{2}
    \left[
         \Vert \boldsymbol{r} \Vert^{2}_{\ccalH}
         -
         2\langle \boldsymbol{r}, F(\boldsymbol{w})\rangle_{\ccalH}
         +
         \Vert F(\boldsymbol{w}) \Vert_{\ccalH}^{2}
    \right]
    .
\end{equation}
Then, it follows that
\begin{multline}
\ell\left(
        \boldsymbol{w}
        +
        \boldsymbol{d}
    \right)
-
\ell\left(
         \boldsymbol{w}
    \right)
    =
    -\left\langle
           \boldsymbol{r}
           ,
           F\left(
                \boldsymbol{w}
                +
                \boldsymbol{d}
            \right)
            -
            F\left(
                \boldsymbol{w}
            \right)
     \right\rangle_{\ccalH}
     +
     \\
     \frac{1}{2}
            \left[
                  \left\langle
                        F\left(
                             \boldsymbol{w}
                             +
                             \boldsymbol{d}
                         \right)
                         ,
                         F\left(
                             \boldsymbol{w}
                             +
                             \boldsymbol{d}
                         \right)
                  \right\rangle_{\ccalH}
                  -
                  \left\langle
                        F\left(
                             \boldsymbol{w}
                         \right)
                         ,
                         F\left(
                             \boldsymbol{w}
                         \right)
                  \right\rangle_{\ccalH}
            \right]
            .
\end{multline}
\begin{multline}
\ell\left(
        \boldsymbol{w}
        +
        \boldsymbol{d}
    \right)
-
\ell\left(
         \boldsymbol{w}
    \right)
    =
    \\
    \left\langle
           F(\boldsymbol{w}+\boldsymbol{d})
            -
           F(\boldsymbol{w})
          ,
          \frac{1}{2}
                \left[
                      F(\boldsymbol{w}+\boldsymbol{d})
                      +
                      F(\boldsymbol{w})
                \right]
                -
                \boldsymbol{r}
    \right\rangle_{\ccalH}
    .
\end{multline}
Now, taking into account the definition of the Fr\'echet derivative of $F$ it follows that
\begin{multline*}
\ell\left(
        \boldsymbol{w}
        +
        \boldsymbol{d}
    \right)
-
\ell\left(
         \boldsymbol{w}
    \right)
    =
    \\
    \left\langle
           \bbD_{F}(\boldsymbol{w})\left\lbrace
                                        \boldsymbol{d}
                                   \right\rbrace
            +
            o\left( \boldsymbol{d} \right)
           ,
           F(\boldsymbol{w})
           +
          \frac{1}{2}
                \bbD_{F}(\boldsymbol{w})
                    \left\lbrace
                         \boldsymbol{d}
                    \right\rbrace
                -
                \boldsymbol{r}
                +
                 o\left( \boldsymbol{d} \right)
    \right\rangle_{\ccalH}
    .
\end{multline*}
Now, we group and rearrange terms again to obtain
\begin{multline}
\ell\left(
        \boldsymbol{w}
        +
        \boldsymbol{d}
    \right)
-
\ell\left(
         \boldsymbol{w}
    \right)
    =
    \left\langle
           \bbD_{F}(\boldsymbol{w})\left\lbrace
                                        \boldsymbol{d}
                                   \right\rbrace
           ,
           F(\boldsymbol{w})
                -
                \boldsymbol{r}
    \right\rangle_{\ccalH}
    +
    \\
     \left\langle
           \bbD_{F}(\boldsymbol{w})\left\lbrace
                                        \boldsymbol{d}
                                   \right\rbrace
           ,
          \frac{1}{2}
                \bbD_{F}(\boldsymbol{w})
                    \left\lbrace
                         \boldsymbol{d}
                    \right\rbrace
    \right\rangle_{\ccalH}
    \\
    +
     \left\langle
           o\left( \boldsymbol{w} \right)
           ,
                     F(\boldsymbol{w})
           +
          \frac{1}{2}
                \bbD_{F}(\boldsymbol{w})
                    \left\lbrace
                         \boldsymbol{d}
                    \right\rbrace
                -
                \boldsymbol{r}
                +
                 o\left( \boldsymbol{d} \right)
    \right\rangle_{\ccalH}
    .
\end{multline}
Then, taking into account the definition of the Fr\'echet derivative, the fact that $o(\boldsymbol{w})\to 0$ and that the term
$
     \left\langle
           \bbD_{F}(\boldsymbol{w})\left\lbrace
                                        \boldsymbol{d}
                                   \right\rbrace
           ,
          \frac{1}{2}
                \bbD_{F}(\boldsymbol{w})
                    \left\lbrace
                         \boldsymbol{d}
                    \right\rbrace
    \right\rangle_{\ccalH}
$
is quadratic with respect to $\boldsymbol{d}$ we have that
\begin{equation}
\bbD_{\ell}(\boldsymbol{w})
     \left\lbrace
           \boldsymbol{d}
     \right\rbrace
     =
     -
     \left\langle
           \bbD_{F}(\boldsymbol{w})\left\lbrace
                                        \boldsymbol{d}
                                   \right\rbrace
           ,
           \boldsymbol{r}
           -
           F(\boldsymbol{w})
    \right\rangle_{\ccalH}
\end{equation}
\end{proof}


\subsection{Proof of Theorem~\ref{thm_Frcht_comp}}
\label{proof_thm_Frcht_comp}

\begin{proof}

Since we know that $F_{1}$ is Fr\'echet differentiable at $\boldsymbol{u}$ we have
\begin{equation}\label{eq_proof_thm_Frcht_comp_1}
F_{2}\left(
       F_{1}\left(
           \boldsymbol{u}
           +
           \boldsymbol{w}
         \right)
       \right)
       =
       F_{2}
       \left(
           F_{1}(\boldsymbol{u})
           +
           \bbD_{F_{1}}(\boldsymbol{u})
                 \{ \boldsymbol{w} \}
                 +
                 o(\boldsymbol{w})
       \right)
       .
\end{equation}
Now, if we take into account the Fr\'echet derivative of $F_{2}\circ F_{1}$ at $\boldsymbol{u}$ we have
\begin{equation}\label{eq_proof_thm_Frcht_comp_2}
F_{2}\left(
       F\left(
           \boldsymbol{u}
           +
           \boldsymbol{w}
        \right)
       \right)
       =
       F_{2}\left(
                 F(\boldsymbol{u})
              \right)
           +
           \bbD_{F_{2}\circ F_{1}}(\boldsymbol{u})
               \{
                  \boldsymbol{w}
               \}
               +
               o(\boldsymbol{w})
           ,
\end{equation}
Then, combining~\eqref{eq_proof_thm_Frcht_comp_1} and~\eqref{eq_proof_thm_Frcht_comp_1} and arranging terms we have
\begin{multline}\label{eq_proof_thm_Frcht_comp_3}
\bbD_{F_{2} \circ F_{1}}(\boldsymbol{u})
    \{
      \boldsymbol{w}
     \}
       =
       \\
              F_{2}
       \left(
           F_{1}(\boldsymbol{u})
            +
            \bbD_{F_{1}}(\boldsymbol{u})
                \{
                  \boldsymbol{w}
                \}
            +
            o(\boldsymbol{w})
       \right)
       -
        F_{2}\left(
                 F_{1}(\boldsymbol{u})
               \right)
        -
        o(\boldsymbol{w})
        .
\end{multline}
Now, we recall that the Fr\'echet derivative of $F_{2}$ is given according to
\begin{equation}\label{eq_proof_thm_Frcht_comp_4}
F_{2}\left(
        F_{1}(\boldsymbol{u})
        +
        \boldsymbol{\xi}
      \right)
      =
F_{2}\left(
        F_{1}(\boldsymbol{u})
      \right)
      +
      \bbD_{F_{2}}\left(
                     F_{1}(\boldsymbol{u})
                  \right)
                    \{
                       \boldsymbol{\xi}
                     \}
      +
      o(\boldsymbol{\xi})
      .
\end{equation}
Then, replacing~\eqref{eq_proof_thm_Frcht_comp_4} in~\eqref{eq_proof_thm_Frcht_comp_3} with $\boldsymbol{\xi} = \bbD_{F_{1}}(\boldsymbol{u})\{ \boldsymbol{w} \} + o(\boldsymbol{w})$ it follows that
\begin{multline}
\bbD_{F_{2} \circ F_{1}}(\boldsymbol{u})
    \{
      \boldsymbol{w}
    \}
=
\bbD_{F_{2}}\left(
               F_{1}(\boldsymbol{u})
            \right)
                    \left\lbrace
                        \bbD_{F_{1}}(\boldsymbol{u})
                             \{
                                \boldsymbol{w}
                              \}
                        +
                        o(\boldsymbol{w})
                    \right\rbrace
                    \\
                    +
                    o\left(
                        \bbD_{F_{1}}(\boldsymbol{u})
                           \{
                             \boldsymbol{w}
                            \}
                            +
                            o(\boldsymbol{w})
                     \right)
                     .
\end{multline}
\end{proof}


\subsection{Proof of Proposition~\ref{prop_derivative_nonlin_and_filter}}
\label{proof_prop_derivative_nonlin_and_filter}

\begin{proof}

Let $F(\boldsymbol{w}) = \boldsymbol{w}\ast f$, then following the definition of the Fr\'echet derivative we have
\begin{equation}
F\left(
         \boldsymbol{w}
         +
         \boldsymbol{d}
     \right)
     -
F\left(
          \boldsymbol{w}
    \right)
    =
    \left(
          \boldsymbol{w}
          +
          \boldsymbol{d}
    \right)\ast f
    -
    \boldsymbol{w}
           \ast
    f
    =
    \boldsymbol{d}
        \ast
    f
    ,
\end{equation}
which implies that $\bbD_{F}(\boldsymbol{w})\{ \boldsymbol{d} \} =  \boldsymbol{d} \ast f$. Now, if we take into account that $\ell (\boldsymbol{w}) = \eta \left( F(\boldsymbol{w})\right)$ and apply the chain rule according to Theorem~\ref{thm_Frcht_comp}, it follows that
\begin{equation}
\bbD_{\ell}(\boldsymbol{w})
         \left\lbrace
              \boldsymbol{d}
         \right\rbrace
         =
         \bbD_{\eta}\left(
                        \boldsymbol{w}
                        \ast
                        f
                    \right)
                    \left\lbrace
                         \boldsymbol{d}
                         \ast
                         f
                    \right\rbrace
                    .
\end{equation}
\end{proof}

\ifCLASSOPTIONcaptionsoff
  \newpage
\fi

\end{document}